# MMUEChange: A Generalized LLM Agent Framework for Intelligent Multi-Modal Urban Environment Change Analysis

Zixuan Xiao[1], Jun Ma[1,*], Siwei Zhang[1]

[1] Department of Urban Planning and Design, The University of Hong Kong, Hong Kong
[*] Corresponding author

**Abstract**

Understanding urban environment change is essential for sustainable development. However, current approaches, particularly remote sensing change detection, often rely on rigid, single-modal analysis. To overcome these limitations, we propose MMUEChange, a multi-modal agent framework that flexibly integrates heterogeneous urban data via a modular toolkit and a core module, Modality Controller for cross- and intra-modal alignment, enabling robust analysis of complex urban change scenarios. Case studies include: a shift toward small, community-focused parks in New York, reflecting local green space efforts; the spread of concentrated water pollution across districts in Hong Kong, pointing to coordinated water management; and a notable decline in open dumpsites in Shenzhen, with contrasting links between nighttime economic activity and waste types, indicating differing urban pressures behind domestic and construction waste. Compared to the best-performing baseline, the MMUEChange agent achieves a 46.7% improvement in task success rate and effectively mitigates hallucination, demonstrating its capacity to support complex urban change analysis tasks with real-world policy implications.

**Keywords:** Change Analysis; Large Language Model (LLM); Multi-modal Agent; Urban Environment

## 1. Introduction

Analyzing urban environment changes is crucial for effective urban planning, sustainable development, and resource management [1,2]. These changes, driven by rapid urbanization and human activities, significantly impact various aspects of urban life [3]. For example, land-use analysis aids urban planners in optimizing resource distribution, while monitoring air and water quality is vital for public health policy formulation [4–6]. Urban environment changes manifest in multiple forms, necessitating customized analytical methods for real-world applications, ranging from descriptive analysis of single factors to multi-level assessments of relationships between environmental variables. For instance, a preliminary analysis might identify areas with significant green space loss, while a more in-depth study could examine the link between this loss and rising temperatures in particular districts. Recognizing these challenges, this study sets out to develop an intelligent framework that facilitates hierarchical, multi-level interpretation of urban environment changes.

The complexity of urban environmental changes is amplified by the multifaceted nature of urban systems. The interactions among factors such as land use, air quality, and population density make it challenging to rely solely on single-modality data or single-method analysis. Effectively addressing this complexity requires fusing multi-modal data (e.g., satellite imagery, sensor data, and social data) and employing a variety of analytical techniques, including deep learning, geo-location alignment, and spatial distribution analysis. Integrating these diverse data types and methods is essential for capturing the full scope of urban environmental dynamics.

Recent studies across diverse contexts further underscore the complexity of urban environmental change. For example, in the Drina River Basin of the Western Balkans, researchers have highlighted how pollution control and ecological protection require integrated, transboundary governance frameworks that transcend administrative boundaries and emphasize cooperative management of shared resources [7]. At the societal level, a 35-year survey in Idaho demonstrated that public perceptions of drinking water quality have declined significantly, with growing reliance on household filtration systems and shifting practices in bottled water consumption [8]. Such findings suggest that urban environmental challenges are not only ecological but also perceptual, shaped by risk awareness and behavioral responses. At the micro-urban scale, case studies from Serampore in the Kolkata Metropolitan Area reveal how the degradation and conversion of ponds under urbanization pressures erode blue infrastructure, reduce ecological resilience, and expose gaps in local policy enforcement [9]. Taken together, these studies illustrate that urban environmental change is inherently multi-scalar, requiring analytical approaches that link regional governance, community perceptions, and localized ecological dynamics.

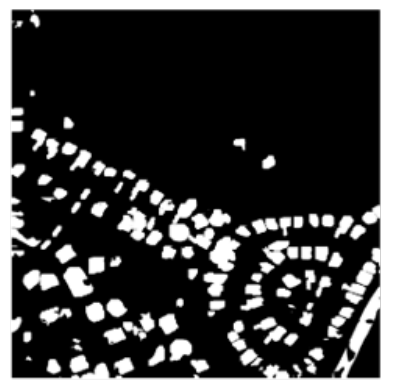

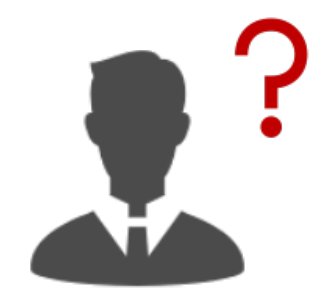

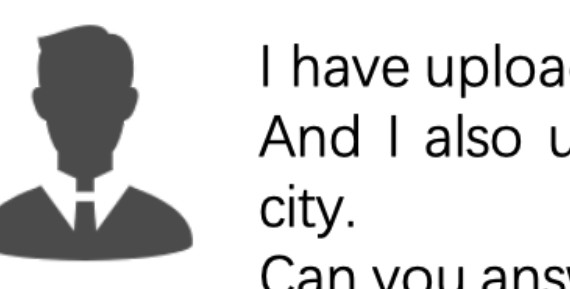

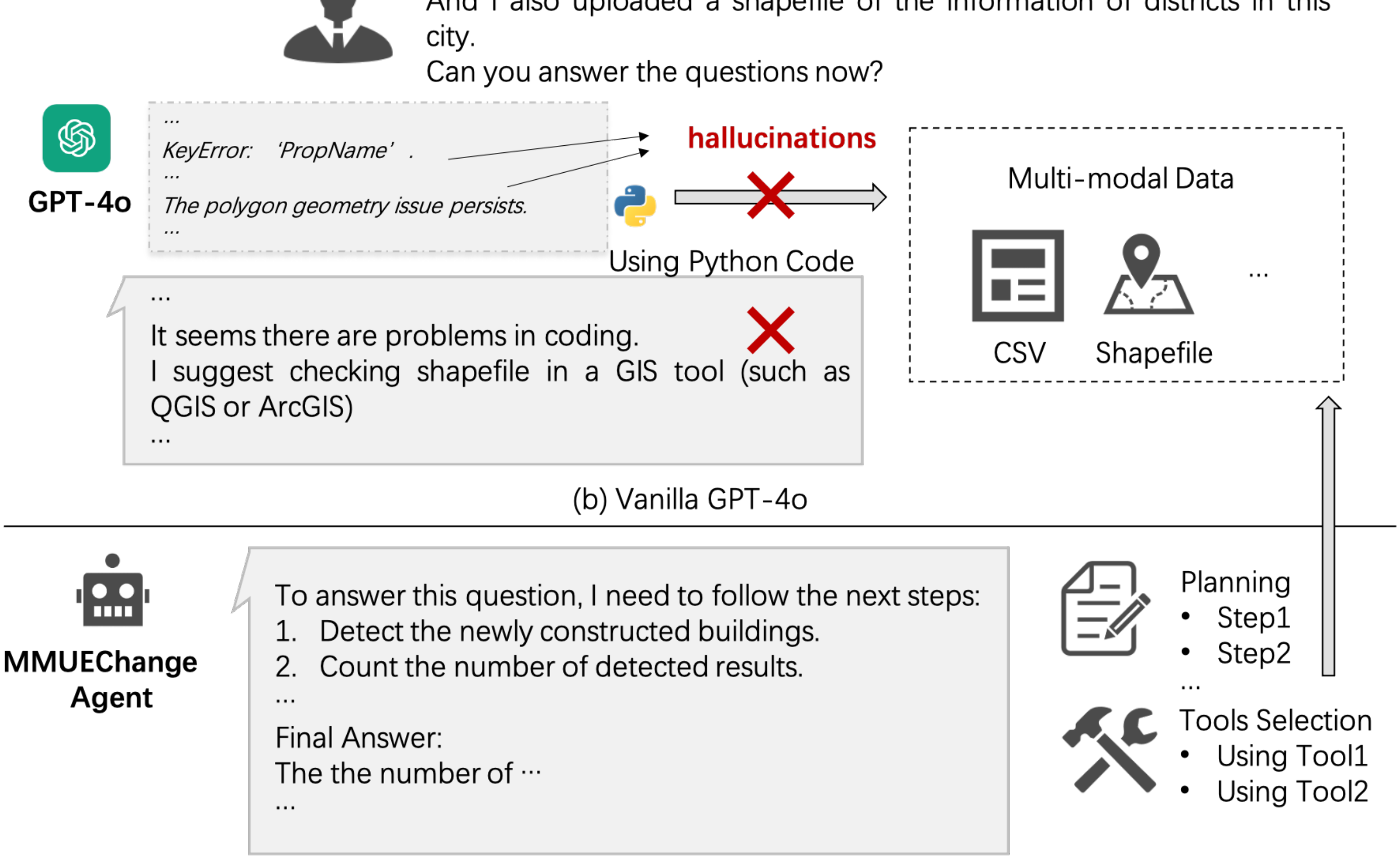

Fig. 1 Comparison of results among the change detection model, Vanilla GPT-4o, and our MMUEChange agent when responding to urban environment change analyzing queries.

(a) The results of the remote sensing change detection model can only provide limited information, unable to respond to the 'what', 'where', and 'why' questions.

(b) The vanilla GPT-4o model attempts to generate Python code to process multi-modal data but fails due to a lack of domain-specific knowledge.

(c) Our MMUEChange agent is capable of planning solution steps and overcoming the hallucination issues by selecting appropriate tools to extract relevant information from multi-modal data.

Remote sensing change detection methods have been foundational in change analysis, particularly for urban environments, by delivering precise pixel-level change maps [10]. However, although these methods deliver highly accurate maps, pixel-level information alone cannot comprehensively capture the complexity of urban environment changes. As illustrated in Fig. 1(a), the challenges of urban environment systems go beyond the isolated detection of pixel-level changes, as they require understanding higher-level patterns or even relationships, such as the ‘what’, ‘where’, and ‘why’ questions, essential for multi-layered analytical needs. This work defines three progressively complex analysis levels: the ‘what’ level, involving basic descriptions, qualitative and quantitative assessments. The ‘where’ level, addressing spatial distributions and their interpretation. The ‘why’ level, exploring influential factors and deeper causal insights. As urban environment change analysis increasingly requires flexible, multi-level insights beyond pixel information, traditional change detection results often fall short, underscoring the need for an intelligent system that supports diverse analytical needs in urban change studies.

Large Language Models (LLMs) have exhibited strong few-shot learning capabilities and have recently advanced intelligent systems, making them vital for multi-modal data fusion in urban environment change analysis [11,12]. However, LLMs are prone to hallucination, a term that refers to the generation of outputs that are syntactically plausible but factually incorrect or unsupported by the input context. Such behavior often arises from limitations in domain-specific knowledge or misalignment between training data and the task at hand, and it directly affects the reliability of LLM-based systems. As shown in Figure 1(b), the vanilla GPT-4o model [13] attempts to construct Python code for multi-modal data processing but fails due to its lack of specialized expertise. This highlights the need for a framework that integrates LLM intelligence with essential domain knowledge to enhance accuracy and effectiveness in urban environment analysis.

In parallel, recent agent frameworks such as LangChain and AutoGen have been developed to extend the raw capabilities of LLMs by connecting them with default tools and enabling structured task orchestration. LangChain offers modular agents, including CSV and pandas DataFrame agents, that facilitate basic interaction with structured data, while AutoGen focuses on multi-agent collaboration through message passing and role specialization. Although effective for general-purpose text or tabular processing, these frameworks remain limited in handling complex domain-specific tasks.

To address these challenges, this study introduces a generalized LLM agent framework tailored for complex urban environment change analysis, supporting multi-modal data fusion and diverse analytical methods. Key to this framework is a modality controller that aligns user queries with corresponding data modalities, paired with a toolkit composed of flexible modules for different data types. As illustrated in Figure 1(c), this intelligent agent effectively plans solutions for various urban issues, selecting appropriate tools, processing data accurately, and generating reliable insights. This framework bridges disparate data sources and analytical methods, enabling a holistic understanding of urban changes to support informed decision-making.

To demonstrate its versatility and adaptability, three case studies in the Experiment Section showcase the application of the framework across diverse urban change scenarios and analytical levels ('what,' 'where,' and 'why'), illustrating how the framework adapts to different application scenarios. Our analysis revealed significant temporal and spatial disparities in park development in New York from 2010 to 2017, with a strategic emphasis on small community parks likely designed to address local green space needs in densely populated areas. In Hong Kong, the agent identified clusters of deteriorating water quality, showing a troubling expansion of poor-quality zones across districts, which suggests a pressing need for coordinated cross-district water management. In Shenzhen, the findings highlighted substantial reductions in open dumpsites, with contrasting influences of nighttime economic activity on waste types: while nighttime activity was positively associated with domestic waste, it showed no significant impact on construction waste. These insights underscore the MMUEChange agent's capability to integrate and interpret diverse data sources, offering policymakers multi-layered urban analysis previously difficult to achieve with traditional models for complex urban environment changes. From another perspective, our experimental results show that the proposed framework offers both generality and flexibility to combine diverse analysis methods for different urban environment change research questions.

In conclusion, this study aims to bridge the gap between high-precision yet limited traditional remote sensing methods and flexible but unreliable vanilla LLMs. By framing urban change analysis within a hierarchical "what–where–why" problem set, the framework directly extends pixel-level change detection toward interpretable spatial and causal insights while constraining LLM reasoning within verifiable and domain-grounded evidence. This layered design ensures both rigor and reliability, addressing the core shortcomings of the two paradigms.

Overall, the contributions of the study can be summarized as follows:

(1) We articulate a hierarchical *"what–where–why"* problem set for urban environment change analysis, which formalizes the levels of description, spatial reasoning, and causal interpretation. This formulation highlights the analytical gaps that traditional remote sensing methods and vanilla LLMs cannot adequately address.

(2) We propose a generalized LLM-agent framework that operationalizes this problem set through multi-modal data alignment and domain grounding.

(3) Based on this framework, we design the MMUEChange agent, which flexibly integrates the modular toolkit to detect key change trends, interpret spatial phenomena, and analyze socio-environmental interactions.

(4) We validate the framework through three real-world case studies, demonstrating reliable, multi-level insights and reduced hallucinations compared with existing approaches.

## 2. Related Work

### 2.1. Change Analysis

Remote sensing-based change detection has long served as a foundational technique for analyzing land cover and land use changes in urban environments. Traditional approaches, including algebra-based, statistics-based, and transformation-based methods, offer computational efficiency but rely heavily on manual thresholding and operate primarily at the pixel level, limiting their interpretability in complex urban contexts [14–16]. With the rise of deep learning, more advanced models such as Post-classification Change Method (PCCM), Direct Classification Method (DCM), Differencing Neural Network Method (DNNM), and recurrent or adversarial architectures have substantially improved change detection accuracy by learning multi-level representations directly from imagery [17–22]. Recent advances further incorporate multi-modal fusion of optical, SAR, and hyperspectral data to enhance map quality [23–25]. Nevertheless, existing methods remain constrained to image-focused analysis and are unable to integrate heterogeneous data types or support multi-level reasoning required for “what,” “where,” and “why” questions, which are essential for contemporary urban environment change analysis studies.

### 2.2. Large Language Models and Multi-modal Agents

Due to the emergence of Large Language Models (LLMs) and their powerful capabilities in natural language understanding and summarization, they have become a crucial component in intelligent systems [13,26]. For harnessing the powerful capabilities of LLMs, prompts are crucial techniques [27]. A prompt is the input provided to an LLM to guide it toward a specific output. Effective prompting is crucial to utilizing these models across various tasks, such as generating text, answering questions, or predicting outcomes. Techniques like Chain of Thought (CoT) and In-Context Learning (ICL) have been developed to enhance the performance of LLMs [28–35]. CoT encourages models to break down complex tasks into smaller, logical steps, while ICL allows LLMs to learn from examples provided in the prompt itself without the need for retraining.

Recent research has further highlighted that the deployment of LLMs is shaped not only by technical prompting strategies but also by broader sociopolitical and infrastructural contexts. For instance, sovereign LLMs developed under national governance frameworks exhibit systematic biases in framing geopolitical narratives, underscoring the need for transparency and cross-border governance mechanisms [36]. In parallel, studies on domain-specific applications in low-resource settings, such as SMS-based agricultural advisory systems in sub-Saharan Africa, demonstrate how LLMs can be adapted through contextual knowledge injection and guardrails to deliver locally relevant and trustworthy outputs [37]. Similarly, in healthcare, pretrained deep learning and NLP models have been integrated into e-health systems to support clinical decision-making, though challenges of scalability, data governance, and adaptability remain [38]. These findings emphasize that ensuring reliability and domain alignment is as crucial as improving prompting techniques when extending LLMs into real-world applications.

As the demand for more complicated intelligent systems grows, multi-modal agents have emerged to integrate various forms of data, such as text, images, audio, and video, within a unified framework [39–49]. These agents often leverage pre-trained encoders that serve as adapters for different modalities, enabling LLMs to process and generate content across multiple forms of modality. For example, systems like SEED-LLaMA and SpeechGPT allow LLMs to comprehend and produce both text and images or audio by converting them into discrete embeddings [50,51]. However, these models require vast amounts of training data across different modalities and considerable computational resources to fine-tune.

Recent agent frameworks (e.g., LangChain, AutoGen) provide modular architectures to orchestrate LLMs with tool invocation, memory, and control logic, but they differ in their design trade-offs. LangChain supports agents such as the CSV agent or Pandas DataFrame agent, where a CSV file is loaded into a DataFrame and exposed to an LLM via a Python REPL tool for queries. AutoGen, on the other hand, emphasizes structured multi-agent collaboration and message passing, with integrations to wrap LangChain tools when needed [52]. While these frameworks are effective for text-centric or tabular tasks, their limitations become evident in scientific multi-modal contexts. LangChain's default agents (e.g., CSV or pandas DataFrame agents) are effective for general tabular operations but lack domain-specific reasoning and cannot seamlessly handle heterogeneous modalities. AutoGen emphasizes multi-agent collaboration, yet its coordination mechanisms introduce overhead and provide limited transparency, making it difficult to trace or adapt reasoning processes. As a result, these frameworks remain insufficient for complex tasks such as multi-modal urban change analysis, where domain alignment, efficient modality fusion, and interpretable decision-making are essential.

Due to the substantial resource demands of training these multi-modal agents, tool-assisted agent construction has emerged as a practical choice [53–55]. Instead of training agents on extensive datasets, this approach connects LLMs with external tools, to perform specific tasks. Some works proposed agents for remote sensing images [56,57], but they lack a specific focus on urban environment change analysis, leading to significant hallucination issues where LLMs generate incorrect or irrelevant information due to a lack of domain-specific knowledge [58]. In contrast, our work introduces a general framework for constructing multi-modal agents for urban environment change analysis. This framework enables the flexible selection of different modules for different data modalities, helping to mitigate the hallucination problem by tailoring the tools as analytical methods for urban change analysis.

## 3. Method

### 3.1. MMUEChange Agent Framework

The proposed framework, as illustrated in Fig. 2, is composed of three main components: a central module referred to as the Modality Controller, an LLM backend, and a modularly designed Toolkit. These components work together to complete intelligent urban environment change analysis, ensuring the adaptability and precision of the framework for diverse queries on multi-modal data.

Table 1 Nomenclature

| Symbol | Definition |
| --- | --- |
| $Q$ | User query in text format. |
| $AL(Q)$ | Analysis level extracted from $Q$; $\in \{what, where, why\}$. |
| $Loc(Q)$ | Location parsed from $Q$. |
| $t(Q)$ | Time or time interval parsed from $Q$. |
| $MC(\cdot)$ | Modality Controller (routing, alignment, aggregation) function. |
| $Q_{aligned}$ | Query after demand alignment. |
| $D$ | Set of available modalities. |
| $D_1/D_2$ | Selected modality data for $Q_{aligned}$; subset of $D$; |
| $D_{aligned}$ | Data after spatial/ID alignment. |
| $L$ | LLM backend. |
| $P$ | Preliminaries and principles. |
| $H_{<h}$ | History before round $h$. |
| $T$ | Modular toolkit. |
| $T_h^i$ | Tool/module selected at round $h$. |
| $R_h^i$ | Result produced for subtask at round $h$. |
| $RP$ | Reasoning and planning by LLM. |
| $Geo-Align(\cdot)$ | Geographical alignment operator. |
| $ID-Align(\cdot)$ | Unique identification connection. |
| $f(\cdot), g(\cdot)$ | Accessors that extract geo-information. |
| $GUID \in U$ | Globally unique identifier from the identifier space $U$. |
| $h, i$ | Indices: interaction round $h$, subtask index $i$. |

For clarity, we follow the notation in Table 1 throughout the Method Section. As the core of the framework, the Modality Controller (denoted as $MC$) performs three critical functions. First, to complete user demand alignment, it identifies the required analysis level ($AL$), location ($Loc$), and time ($t$) dimensions through three sub-processes: change analysis level deciding, site locating, and time confirming, ensuring that the input query ($Q$) is precisely interpreted. Mathematically, this can be represented as:

$$Q_{aligned} = MC(AL(Q), Loc(Q), t(Q)) \tag{1}$$

where $AL(Q)$ denotes the analysis level of the query, $Loc(Q)$ represents the identified location and $T(Q)$ corresponds to the time of the query.

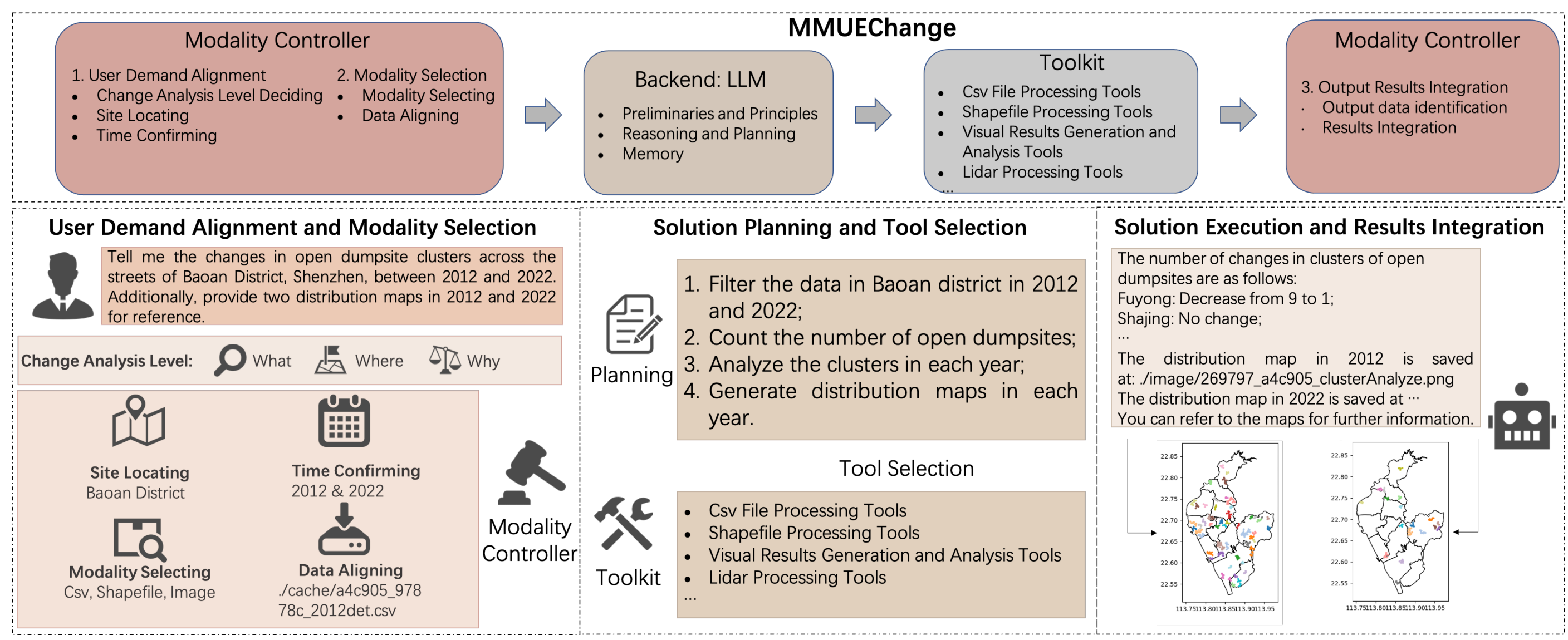

Fig. 2 The framework is composed of three component: Modality Controller, LLM, and Toolkit.

Second, after user demand alignment is complete, it selects the appropriate modality data ($D_{mod}$) based on $Q_{aligned}$, represented as:

$$D_{mod} = MC(Q_{aligned}) \tag{2}$$

Next, the selected modality data ($D_{mod}$) aligns the selected modality with locally stored multi-modal data by assigning globally unique identifier ($GUID$), ensuring that subsequent processing operations interact with the correct data. Once this relevant background information is obtained, the Modality Controller synchronizes the information with the LLM backend.

The LLM Backend ($L$) provides the necessary support for the framework consisting of three subcomponents: preliminaries and principles ($P$), reasoning and planning ($RP$), and memory ($H$). Since most LLMs are trained on large, general datasets, it is crucial to establish a specialized profile and standardize input and output formats by defining the preliminaries and principles, ensuring the output aligns with the intended objectives. When managing complex tasks, reasoning and planning decomposes the task into sub-tasks, enhancing whole task completion success rates, expressed as:

$$R_h = \{R_h^1, R_h^2, \dots, R_h^i, \dots\} = L(P, Q_{aligned}) \tag{3}$$

In subsequent processing, each sub-task at the $h-th$ round, specifically the $i-th$ sub-task, is handled based on the accumulated history of queries and responses ($H_{<h}$) and prior results ($R_{<h}^i$). This iterative process can be represented as:

$$R_h^i = L(H_{<h}, R_{<h}^i, T_h^i\left(D_{aligned}{}_h^i\right)) \tag{4}$$

where $H_{<h}$ is collection of queries and responses before round $h$ and $T_h^i\left(D_{aligned}{}_h^i\right)$ is the pair including data and corresponding toolkit module.

$$H_{<h} = \{(Q_1, A_1), (Q_2, A_2), \ldots, (Q_{h-1}, A_{h-1})\} \tag{5}$$

$$T_h = \{T_h^1, T_h^2, \ldots, T_h^i, \ldots\} \tag{6}$$

This memory design enables the model to maintain coherent, multi-turn dialogues. The modular toolkit ($T$) is designed to handle the varied modality data ($D_{aligned}$) encountered in urban environment change scenarios. Each modality is paired with a module containing corresponding processing tools ($T_h^i$). Finally, after completing all the intermediate processes ($\{R_h^1, R_h^2, \ldots, R_h^i, \ldots\}$), the Modality Controller uniquely identifies the generated data and aggregates the output results, delivering an accurate response to the input query.

$$A_h = MC(R_h) = MC(\{R_h^1, R_h^2, \ldots, R_h^i, \ldots\}) \tag{7}$$

This modular design allows for precise retrieval and fusion of information from multi-modal data, tailored to the specific needs of each task.

The framework provides flexibility in handling heterogeneous data, allowing for comprehensive analysis across different domains of urban environment change. Additionally, it improves the accuracy and relevance of results by aligning multi-modal data and processing queries systematically. From another perspective, the modular design ensures scalability and adaptability, enabling the framework to manage increasingly complex tasks without major modification, thus advancing research in urban environment analysis in the future. By introducing this generalized framework, we constructed our MMUEChange agent capable of integrating data across various modalities and applying diverse analytical methods, addressing the wide range of urban environment change applications.

### 3.2. Modality Controller and Modular Toolkit

This section provides a detailed introduction of the core module within the framework: the Modality Controller. As previously mentioned, the Modality Controller performs three primary functions: user demand alignment, modality selection, and output results integration. In particular, when aligning with user demand, the Modality Controller first decides the change analysis level, which categorizes the query into three hierarchical levels:

'what', 'where', and 'why'. This leveled approach allows for increasingly sophisticated analysis, supporting a comprehensive investigation of urban environment changes.

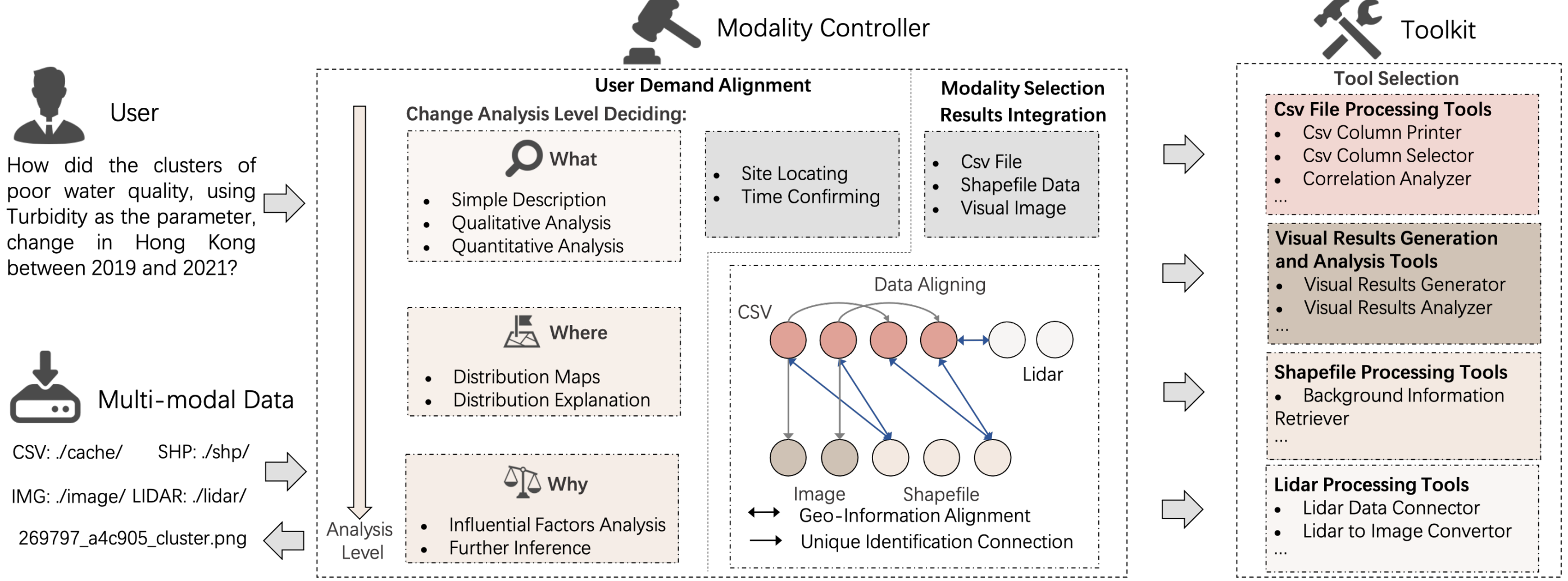

Fig. 3 The detailed workflow of the Modality Controller and Toolkit.

Following user demand alignment ($Q_{aligned}$), modality selection and data alignment are critical components of the framework. Precise data alignment between modalities ensures consistency and reliability in the analysis process. There are two primary alignment methods. The first is geo-information alignment ($Geo - Align$), which relies on geographical metadata, such as latitude and longitude information, to align data between different sources. This alignment can be represented as:

$$D_{aligned} = Geo - Align(f(D_1), g(D_2)) \tag{8}$$

Here $f$ and $g$ are functions of geo-information. For instance, a CSV file containing spatial coordinates $f(D_1)$ can be aligned with a shapefile ($g(D_2)$). This method ensures consistency across multi-modal data, allowing for accurate geographic-based analysis. The second method is unique identification connection ($ID - Align$), used to maintain the relationship between data before and after processing. By assigning a globally unique identifier ($GUID$) to each data, the system tracks parent-child relationships between original and processed data, ensuring traceability. This alignment be expressed as:

$$D_{aligned} = ID - Align(GUID_1, GUID_2) \tag{9}$$

This enables the system to preserve the consistency of data across processes, making it possible to monitor the entire data lineage and facilitating precise information fusion. These two alignment designs enhance the capacity to manage complex data transformations and multi-modal information fusion of our MMUEChange agent.

Table 2 Technical Foundations and Representative Tools of the Modular Toolkit

| Data Modality | Representative Tools | Technical Details | Example Use in Case Studies |
|---|---|---|---|
| **CSV** | Column selection, temporal filtering, string/numeric screening, tabular joins, descriptive statistics | Implemented with **pandas** for efficient tabular processing, supporting flexible filtering, aggregation, and joining operations | Extract and filter NYC park records (2010–2020) to analyze temporal trends in green space expansion |
| **Shapefile** | Projection handling, geometry validation, spatial overlay and join | Based on **geopandas** and **shapely**, enabling coordinate system harmonization, topology repair, and fine-grained spatial queries | Integrating Hong Kong district boundary data with water quality measurements to examine the spatial heterogeneity of urban aquatic conditions |
| **LiDAR** | Area-of-interest clipping, land-cover rendering, class-based summarization | Utilizes **rasterio** and **laspy** for raster/point cloud operations, including tiling, clipping, and semantic rendering | Extracting LULC information from LiDAR-derived surface models to quantify structural transformations in the urban environment |
| **Image Visualization** | Heatmap construction, contour extraction, thematic mapping | Built upon **matplotlib** and **geopandas** visualization layers, allowing high-resolution heatmaps, spatial contouring, and map overlays | Generate urban waste hotspot maps and water quality heatmaps for visual identification of critical zones |
| **Analytical Add-ons** | Clustering, change aggregation, factor attribution | Employs **scikit-learn** for clustering (e.g., DBSCAN), regression-based attribution, and correlation analysis | Conducting correlation analysis between socio-economic and spatial indicators and variations in open dumpsite counts to infer potential urban development drivers of waste generation and reduction |

Once the modality is selected and the data is properly aligned, the corresponding modular tools are invoked to process the data specific to that modality. Each module contains a set of tools tailored to the distinctive characteristics of its data type, ensuring that the appropriate operations are systematically applied. Table 2 provides a structured synthesis of this modular toolkit, outlining representative tools, their technical foundations, and case-specific applications. This modular design significantly enhances scalability, as new tools or modalities can be easily integrated into the framework, enabling diverse analytical methods to adapt to a wide range of urban environment change scenarios. In practice, extending the toolkit involves encapsulating new operations into modular components, such as adding traffic flow parsers or social media text miners, which can then be registered within the modality controller. This plug-and-play process allows the framework to accommodate emerging data types with minimal disruption to the existing pipeline. The modular structure also improves efficiency, as each tool is optimized for its modality, resulting in faster and more accurate data processing. Meanwhile, with specific tools, our agent significantly mitigates the hallucination issues faced by LLMs.

## 4. Experiment

### 4.1. Experiment Setting

To operationalize the hierarchical "what–where–why" analytical paradigm introduced earlier, we design three case studies that correspond to progressively deeper levels of urban environment change analysis. The first case study focuses on descriptive "what" questions related to newly constructed parks, with the ten questions covering basic identification, qualitative interpretation, and quantitative measurement of observed changes. The second case study, centered on coastal water quality, extends the scope to both "what" and "where" analyses. Its questions require numerical assessment of turbidity indicators as well as the generation and interpretation of spatial distribution patterns. The third case study integrates all three levels, "what," "where," and "why", to investigate changes in urban open dumpsites. Here, the questions progress from quantitative summaries to spatial clustering analyses and, finally, to inferential tasks that explore potential drivers based on multi-source contextual data. Across all three cases, we construct ten questions tailored to the analytical depth of each scenario, allowing for a structured and consistent evaluation of the agent's capabilities in multi-modal reasoning and urban environment change interpretation.

Table 3 The question datasets used in case studies.

| Case Study | Analysis Level | Subtype | Number | Example |
|---|---|---|---|---|
| 1 | What | Basic | 4 | Can you tell me the names of the parks constructed in 2015 that are smaller than 2 acres? |
| | | Qualitative | 2 | Please provide the names, addresses, and boroughs of the parks constructed in 2015, along with the number and locations of drinking fountains inside the parks, if any |
| | | Quantitative | 4 | Please provide the names of the parks constructed in December 2017 and the land cover type proportions of their sites as they were in 2010. |
| 2 | What | Quantitative | 6 | Please tell me the average change trends in water quality, using Turbidity as the parameter, across different regions of Hong Kong between 2016 and 2021. |
| | Where | Distribution Maps | 2 | Can you give me a distribution map of clusters with poor water quality, using Turbidity as the parameter, in Hong Kong in 2021? |
| | | Distribution Explanation | 2 | Which district in Hong Kong had the most medium or large clusters of poor water quality, using Turbidity as the parameter, in 2021? |
| 3 | What | Quantitative | 4 | Could you provide the number and changes in open dumpsites for the different streets in Baoan District, Shenzhen, in 2012 and 2022? |
| | Where | Distribution Maps | 2 | Could you provide me with a distribution map of open dumpsite clusters across the streets of Baoan District, Shenzhen, in 2022? |
| | | Distribution Explanation | 2 | How did the number of open dumpsite clusters change across the streets of Baoan District, Shenzhen, in 2012 and 2022? |
| | Why | Influential Factors Analysis | 2 | Please explain the correlation between the changes in the number of open dumpsites and the population changes across the streets of Baoan District, Shenzhen, between 2012 and 2022. Additionally, provide further analysis based on the results. |

The construction of these question datasets ensures a systematic evaluation of each case study by providing a structured evaluation from basic to advanced analysis. Moreover, the hierarchical construction promotes a deeper understanding of urban environment change through each level of analysis, what, where, and why. By structuring the experiment in this manner, these three case studies not only showcase the wide-ranging applicability of the proposed framework but also illustrate its adaptability to different scales and types of urban environment change problems. Moreover, they validate the capabilities of large language models (LLMs) as powerful tools for urban environment research, highlighting their utility in analyzing, synthesizing, and interpreting multi-modal datasets and different analytical methods across diverse urban contexts.

The three cities were selected deliberately to demonstrate the adaptability of the proposed framework to distinct real-world challenges. New York represents a developed urban context where long-term urban planning and equitable access to green spaces are central issues. Hong Kong exemplifies a coastal metropolitan environment where monitoring water quality is of critical importance. As a major harbor city, its coastal waters are subject to intense anthropogenic pressures, and water quality degradation directly affects public health, marine ecosystems, and cross-district governance of shared coastal resources. Shenzhen, as a rapidly urbanizing city, highlights the governance challenges of waste management under fast economic growth and evolving social practices. Together, these cases capture different governance contexts, environmental domains, and urbanization trajectories, thereby validating the framework's capacity to adapt across heterogeneous urban scenarios.

In our implementation, the MMUEChange agent employs GPT-4o (temperature fixed at 0) as the central reasoning backend LLM. The prompting schema is structured and rule-based, consisting of role definition, a canonical tool-use protocol, and grounding constraints. A detailed introduction of this schema is provided in Appendix A, which illustrates how structured prompting governs the interaction between the LLM and the modular toolkit.

### 4.2. Case study 1: Newly Constructed Urban Parks Description in New York

#### 4.2.1. Data and Workflow Introduction

The analysis of newly constructed parks plays a critical role in understanding urban environment change, as these green spaces directly impact land use, ecological dynamics, and public well-being. Moreover, they alter land cover patterns, replacing built-up or underutilized areas with green, recreational spaces. Studying the transformation of LULC through newly constructed parks provides valuable insights into how urban environments evolve and how such changes align with broader sustainability and livability goals.

The data used in this study comes from NYC Open Data, including three CSV files [59-61] and two LiDAR datasets [62, 63], providing comprehensive information about parks and their surrounding environments. Specifically, the CSV files consist of:

1. Park Property Table: a dataset that contains the basic details of parks in New York, including attributes such as park names, locations, and establishment dates.
2. Public Structures Table: this file describes public structures within parks, such as restrooms, playgrounds, and other amenities that contribute to park usability.
3. Drinking Fountains Table: This dataset specifically lists the locations and conditions of drinking fountains in the parks, providing detailed information on the infrastructural elements within the parks.

All three CSV files are interconnected through a unique foreign key called 'GISPROPNUM', which enables cross-referencing and integration of information across the different datasets. This relationship is important for aligning the features of each park with its public structures and drinking fountains, enabling both qualitative and quantitative analysis of these parks.

The LiDAR data provides highly detailed LULC information for New York City in two time periods: 2010 and 2017. These datasets are used to examine changes in land use patterns over time, particularly focusing on how newly constructed parks have altered the land cover. The LiDAR data allows for accurate quantification of different land cover types, such as grass, tree canopy, and built environments.

As illustrated in Figure 4, the overall workflow involves several steps. After user demand alignment, the agent identifies that there are three levels of queries:

1. Basic Information Query: Identify park names constructed after 2015.
2. Qualitative Analysis: The second step delves into the public structures within these parks, as described in the Public Structures Table. This qualitative analysis helps assess how these structures contribute to park functionality and user engagement.

3. Quantitative Analysis: The final step involves analyzing the land use and land cover changes between 2010 and 2017. The agent needs to compare the two LiDAR datasets to calculate changes in the proportions of land cover types. This quantitative analysis reveals how park development has reshaped the urban landscape.

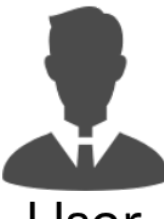

Please provide the names of the parks constructed after 2015 and the structures in them (if any). What' s more, tell me the land cover type proportions of their sites as they were in 2010 and 2017.

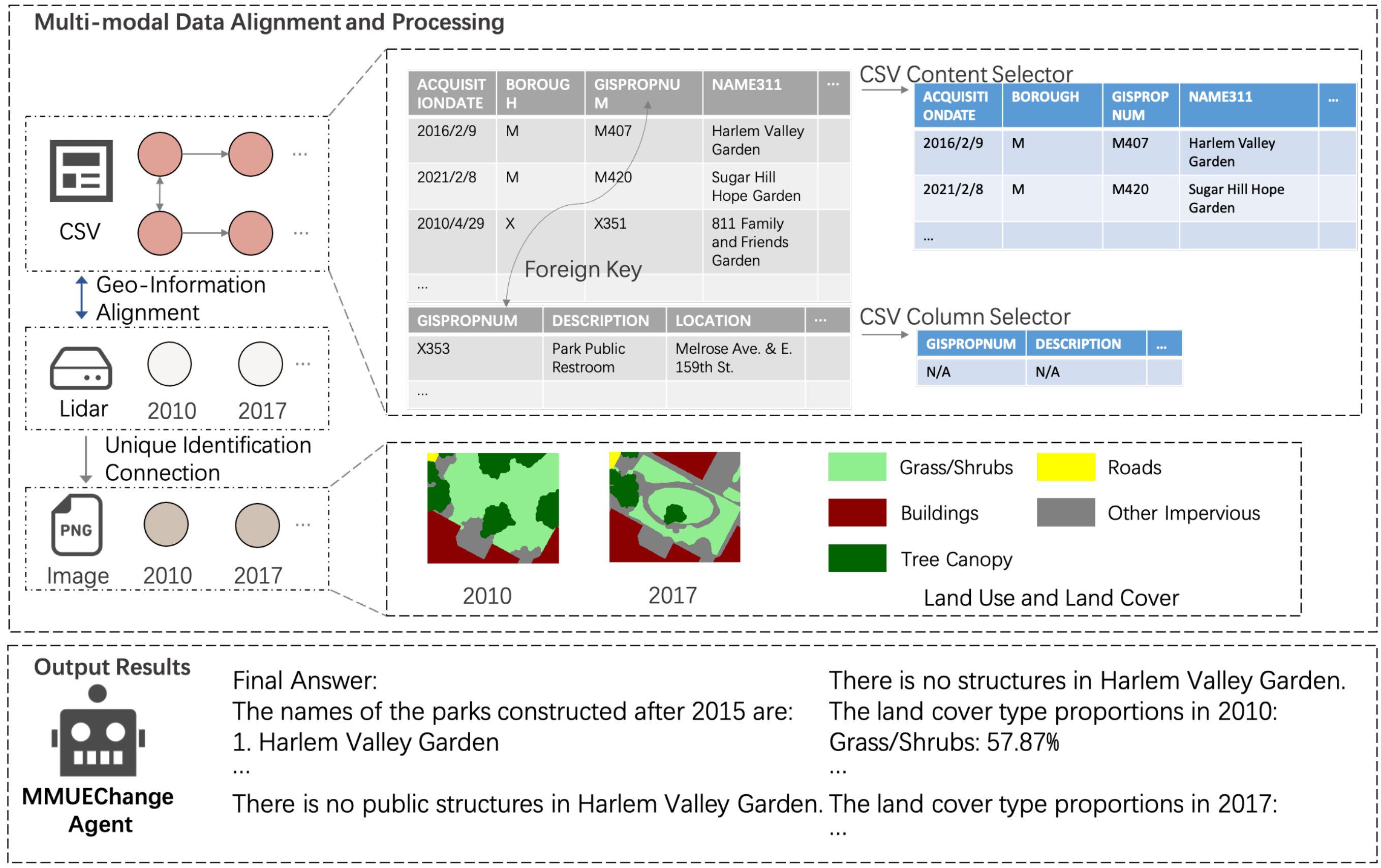

Fig. 4 The overall workflow of case study 1.

After the agent performs extensive data alignment, multi-modal data processing by tools is completed, followed by planning intermediate steps, to provide accurate answers. The original CSV datasets are aligned using the 'GISPROPNUM' foreign key, integrating basic park information and public structures. The alignment between the CSV and LiDAR datasets is achieved through geo-information aligning. Specifically, park coordinates in the CSV files are matched with land cover data in the LiDAR files, ensuring that park-related changes are accurately reflected in the spatial data. After the agent utilizes the tool to process the data, the LiDAR datasets and the generated image files that are used for the calculation of LULC proportions are linked with unique identifications. Finally, the agents provide precise responses through output results integration.

#### 4.2.2. Experimental Performance

As mentioned in Section 4.1, we evaluated the MMUEChange agent using the constructed question dataset to assess its ability to answer questions across varying levels of complexity. To better understand the contribution of individual components, we compared MMUEChange against five baseline agents, each representing a different ablated configuration of the full framework:

- LangChain Default Agent: An agent that is provided with the relevant data sources and leverages LangChain's built-in tools (e.g., SQL agent, pandas DataFrame agent, CSV agent) to process structured data [64]. While this setup enables basic access and manipulation of tabular information, the tool capabilities are general-purpose and lack domain-specific precision or multi-modal coordination. This baseline reflects a scenario where data is available, but the analytical tools are insufficiently tailored to extract deeper insights or handle complex change patterns.
- Standalone Agent: An agent that solely utilizes the LLM backend model without access to external data or toolkit modules. It relies entirely on parametric knowledge and demonstrates the limitations of LLM-only reasoning in real-world tasks.
- No-alignment Agent: This variant is equipped with the full set of modular tools spanning multiple modalities, but excludes the Modality Controller, a key component mainly responsible for aligning data sources both across different modalities and within the same modality across processing stages (e.g., raw vs. derived data). By disabling this controller, the agent lacks structured coordination among heterogeneous inputs, allowing us to evaluate the critical role of data alignment in enabling coherent and accurate multi-step reasoning.
- Data-only Agent: This version incorporates raw data files (e.g., CSVs) but does not equip the agent with any specialized modular tools for structured data handling. It serves to isolate the effect of data availability without targeted processing capabilities.

- Single-modality Agent: An agent equipped with both data access and a toolkit restricted to a single modality (e.g., only CSV tools). This configuration tests the effectiveness of partial tool support in the absence of multi-modal information integration.

The results are presented in Table 4. The LangChain SQL agent, pandas DataFrame agent, and CSV agent, while effective at retrieving structured data, faced limitations when dealing with larger datasets. The size of the CSV files often exceeded the context window limits of the backend LLM, leading to incomplete data loading or failure to accurately answer the questions. Additionally, these agents sometimes modified the original data inadvertently, resulting in data contamination and incorrect outputs. Although LangChain default agents offer a straightforward way to construct agents for handling structured data, whether in the form of SQL databases, pandas dataframes, or CSV files, they are inherently limited by their rigid architecture and default tools. This lack of flexibility makes them ill-suited for more specific or complex applications, especially when the tasks require dynamic data integration or multi-modal processing.

The Standalone Agent operates entirely without external data or tools. As a result, its answers rely solely on the LLM's internal knowledge, often leading to hallucinations and fabricated content. The No-alignment Agent, though equipped with the full toolkit and data, lacks the Modality Controller, which is crucial for aligning data both across modalities and within single modalities across preprocessing stages. This misalignment leads to incorrect file references, broken relational joins, and analytical errors during reasoning, as the agent fails to establish a consistent data schema. The Data-only Agent successfully addressed the basic sublevel questions by accurately retrieving data from a single CSV file. However, it struggled with the qualitative and quantitative analysis sublevels, as it could not precisely identify the connecting foreign key or handle multi-modal data integration.

Table 4 Comparison of performance of our MMUEChange Agent and other baseline models in case study 1.

| Change Analysis Level | What | | | | | | | | | | Accuracy (Correct / Total) |
|---|---|---|---|---|---|---|---|---|---|---|---|
| Subtype | Basic | | | | Qualitative | | Quantitative | | | | / |
| Models \ Questions | Q1 | Q2 | Q3 | Q4 | Q5 | Q6 | Q7 | Q8 | Q9 | Q10 | / |
| LangChain Default Agent | × | × | × | × | × | × | × | × | × | × | 0/10 |
| Standalone Agent | × | × | × | × | × | × | × | × | × | × | 0/10 |

| | | | | | | | | | | |
|---|---|---|---|---|---|---|---|---|---|---|
| No-alignment Agent | ✗ | ✗ | ✗ | ✗ | ✗ | ✗ | ✗ | ✗ | ✗ | ✗ | 0/10 |
| Data-only Agent | ✓ | ✓ | ✓ | ✓ | ✗ | ✗ | ✗ | ✗ | ✗ | ✗ | 4/10 |
| Single-modality Agent | ✓ | ✓ | ✓ | ✓ | ✓ | ✓ | ✗ | ✗ | ✗ | ✗ | 6/10 |
| MMUEChange Agent | ✓ | ✓ | ✓ | ✓ | ✓ | ✓ | ✓ | ✓ | ✓ | ✓ | 10/10 |

The Single-modality Agent, which supported only CSV data through relevant tools, performed better with basic and qualitative analysis questions. It accurately linked multiple CSV files and answered queries involving more complex processing steps. However, due to lack of ability for multi-modal data alignment, it was unable to handle multi-modal data integration, such as correlating CSV data with LiDAR files, and therefore failed to answer the quantitative analysis questions that required spatial and temporal analysis using both data types.

In contrast, the MMUEChange agent demonstrated good performance across all three subtypes of analysis. It accurately answered the basic, qualitative, and quantitative analysis questions, thanks to its advanced data alignment and processing capabilities. Specifically, it successfully aligned CSV data through the 'GISPROPNUM' foreign key, linked CSV and LiDAR datasets using geo-information matching, and ensured the correct connection between LiDAR datasets and image files through a unique identification system. This allowed the agent to handle complex questions involving multi-modal data, such as comparing land use changes for certain parks between 2010 and 2017, something that other agents could not complete.

#### 4.2.3. Key Findings Discussion

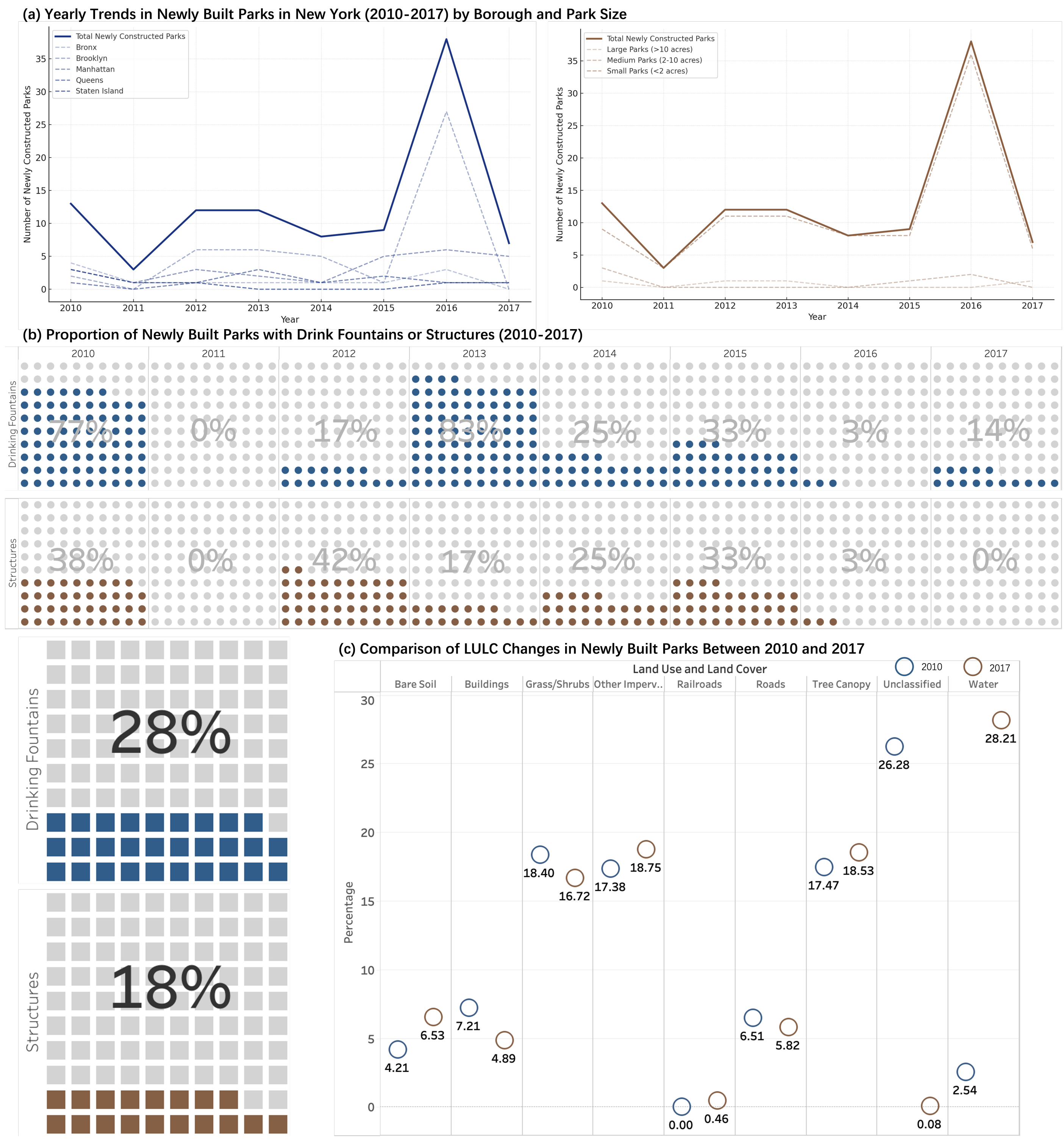

Fig. 5 Analysis of Changes in Newly Constructed Parks in New York (2010-2017).

After utilizing the proposed MMUEChange agent with the question dataset, we obtained valuable insights into the development of newly built parks in New York City between 2010 and 2017. The information derived from the agent allows us to analyze the patterns and characteristics of these new parks across different boroughs and years, facilitating a deeper understanding of urban park expansion trends.

Figure 5 (a) reveals significant temporal and spatial disparities in park development from 2010 to 2017. The yearly trends indicate a peak in newly constructed parks in 2016, which stands in contrast to the noticeably lower numbers in 2011. This uneven distribution suggests a varying prioritization of park development over time, potentially driven by policy shifts, budget allocations, or changing urban needs. A closer examination of borough-specific patterns shows that Brooklyn experienced the most substantial growth in new parks compared to other boroughs. Furthermore, when classified by size, the data shows a predominant increase in smaller parks (<2 acres), while medium and large parks were added at a more consistent rate. This suggests a strategic preference for smaller, community-oriented green spaces that may be easier to implement and manage within densely populated areas.

Figure 5 (b) highlights interesting trends in the amenities included in newly built parks. Specifically, the proportion of parks equipped with drinking fountains peaked in 2010, 2013, and 2015, while structures were more frequently included in parks built in 2010, 2012, and 2015. Conversely, in 2011 and 2016, a considerable proportion of new parks lacked both drinking fountains and structures. These findings indicate variability in park design priorities, which could be attributed to shifts in public preferences, budgetary constraints, or changes in policy focus. The overall proportion of parks with drinking fountains (28%) and structures (18%) remains relatively low, reflecting a potential emphasis on essential green spaces over fully equipped recreational areas. This trend raises questions about the balance between maximizing green space and providing essential amenities, which could impact public satisfaction and park utilization.

Figure 5 (c) demonstrates the influence of new park development on LULC changes. Notably, the proportion of unclassified land significantly decreased, which likely indicates the conversion of previously underutilized or undeveloped areas into structured green spaces. Concurrently, there was a marked increase in water coverage, possibly due to the creation of water bodies such as ponds or artificial lakes within the new parks. This trend could suggest a deliberate strategy to enhance biodiversity, improve aesthetics, or support stormwater management initiatives. The cumulative effect of these changes hints at a multifaceted approach to urban park development, which not only expands recreational areas but also enhances ecological and environmental resilience.

Overall, the results from these analyses provide a comprehensive understanding of the characteristics and impacts of newly constructed parks in New York. Moreover, the experiment highlights the limitations of existing agents in multi-modal urban environment change analysis. In contrast, our MMUEChange agent successfully managed all the experimental queries, demonstrating its ability to address the challenges through extracting information from multi-modal data by different analytical methods in urban environment research.

### 4.3. Case study 2: Coastal Water Quality Change Monitoring in Hong Kong

#### 4.3.1. Data and Workflow Introduction

Monitoring coastal water quality is important for understanding and managing the health of marine ecosystems, particularly in urban areas where industrial and human activities can significantly affect water conditions. Effective water quality monitoring not only helps in detecting pollution hotspots but also supports long-term sustainability efforts by providing data to guide environmental protection policies. For this case study, we first conduct data pre-processing. The primary dataset is composed of in situ water quality data provided by the Hong Kong Environment Protection Department [65]. Although this dataset provides valuable direct measurements, the spatial sparsity of in situ data points is a major limitation. To overcome this issue, we incorporate remote sensing data from the Landsat 8 Collection 2 Provisional Aquatic Reflectance Product [66] to enhance the spatial coverage and make denser water quality predictions. We trained the XGBoost [67, 68] model with the data to predict water quality parameters, such as turbidity, across a larger and more continuous spatial granularity. Finally, densely predicted water quality data is stored in a CSV file for further analysis and visualization.

As mentioned before, the question dataset for this case study is structured around two change analysis levels: ‘what’ and ‘where’. These questions require the agent not only to generate visual outputs, such as maps that highlight the distribution of water quality parameters, but also to interpret these results, identifying areas of concern, such as heatpoint clusters where water quality is particularly poor and concentrated.

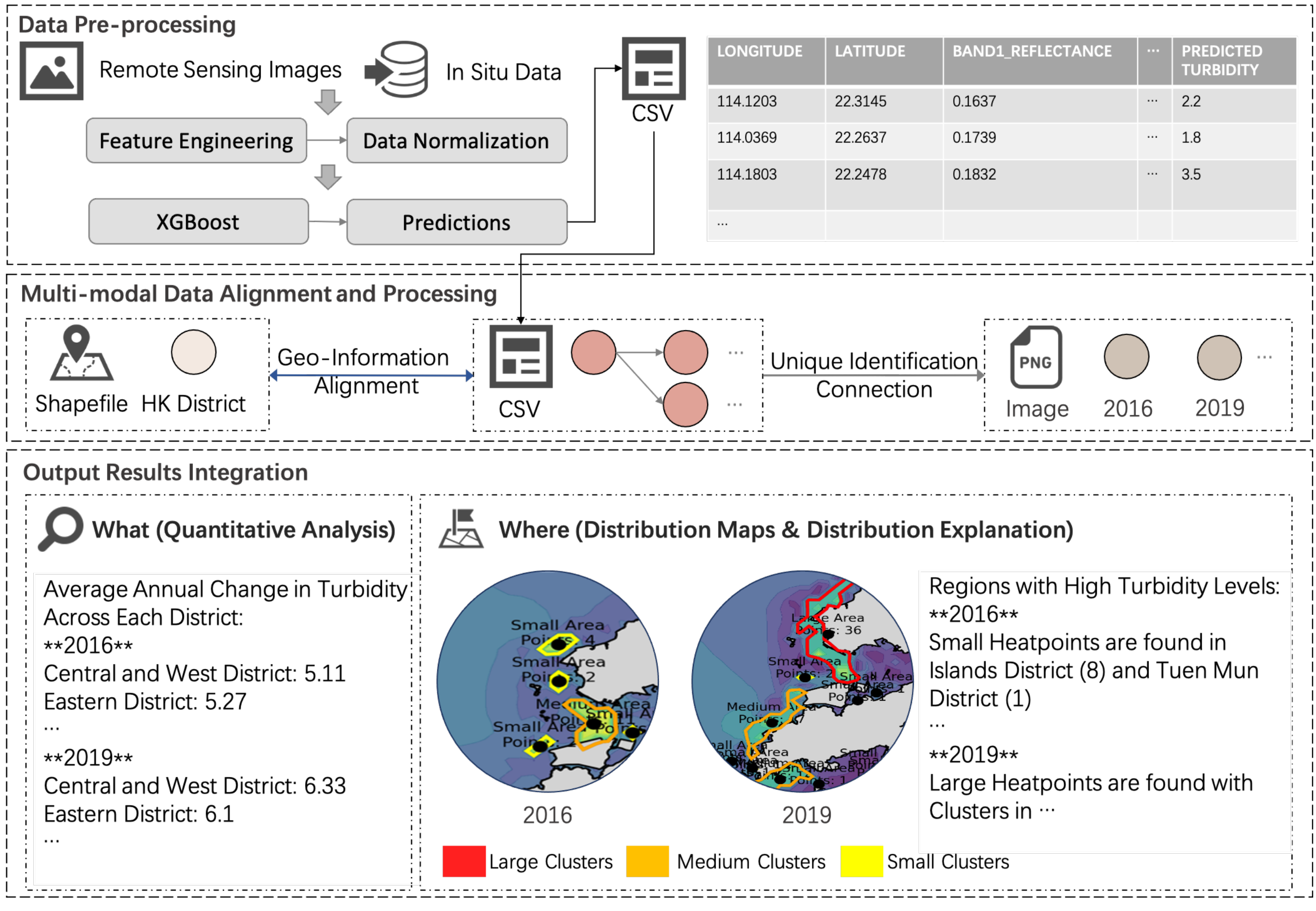

Fig. 6 The overall workflow in case study 2.

As illustrated in Figure 6, the 'what' level analysis leverages the dense water quality predictions stored in the CSV file and a district shapefile [69] for Hong Kong. This allows the agent to calculate and summarize district-level change statistics, such as average water quality change across various districts. At the 'where' level, the agent should identify and visualize areas with concentrated poor water quality (heatpoint clusters) in different years. For example, it generates maps that visually depict these clusters and highlights the districts with the most significant issues. Additionally, it provides a textual interpretation of the maps and their changing trends.

### 4.3.2. Experimental Performance

To evaluate the performance of the MMUEChange agent in the water quality change scenario, we compared it with several baselines: Standalone Agent, No-alignment Agent, Data-only Agent, and Single-modality Agent. The results, presented in Table 5, demonstrate that the Standalone Agent, lacking both data and tools, frequently produced hallucinated answers. The No-alignment Agent, though equipped with full data and tools, often failed to match pre- and post-processed records across modalities, resulting in incorrect tool calls and inconsistent outputs. The Data-only Agent struggles significantly with this task, which is primarily caused by the issues of hallucination when handling multi-modal data such as shapefiles. The Single-modality Agent performs better in terms of answering text-based questions at the 'what' level, but its inability to handle visual outputs limits its effectiveness. Since the agent cannot generate or interpret visual results, it is unable to address any of the 'where' level questions that require spatial analysis and visual representation of water quality change data.

Table 5 Comparison of performance of our MMUEChange Agent and other baseline models in case study 2.

| Change Analysis Level | What | | | | | | Where | | | | Accuracy (Correct / Total) |
|---|---|---|---|---|---|---|---|---|---|---|---|
| Subtype | Quantitative | | | | | | Distribution Maps | | Distribution Explanation | | / |
| Models \ Questions | Q1 | Q2 | Q3 | Q4 | Q5 | Q6 | Q7 | Q8 | Q9 | Q10 | / |
| Standalone Agent | × | × | × | × | × | × | × | × | × | × | 0/10 |
| No-alignment Agent | × | × | × | × | × | × | × | × | × | × | 0/10 |
| Data-only Agent | × | × | × | × | × | × | × | × | × | × | 0/10 |
| Single-modality Agent | ✓ | ✓ | ✓ | ✓ | ✓ | ✓ | × | × | × | × | 6/10 |
| MMUEChange Agent | ✓ | ✓ | ✓ | ✓ | ✓ | ✓ | ✓ | ✓ | ✓ | ✓ | 10/10 |

In contrast, our MMUEChange agent demonstrates good performance across both levels of analysis. It accurately handles both the quantitative and distribution-related tasks, integrating the CSV water quality data with the district shapefile to generate insightful visual outputs. The agent is not only capable of generating clear and precise maps of water quality distribution but also able to interpret these visuals, providing meaningful explanations of the spatial patterns between different years.

#### 4.3.3. Key Findings Discussion

Similarly, after querying the MMUEChange agent using the specified question dataset, we were able to gather a range of insights regarding coastal water quality changes in Hong Kong. These findings allow us to investigate patterns and trends in district-level water quality over a span of several years, highlighting key areas of concern and shifts in water quality parameters.

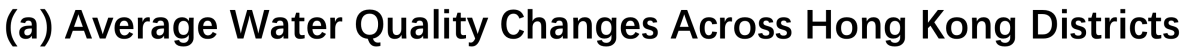

(b) Changes in the Number of Heatpoint Clusters

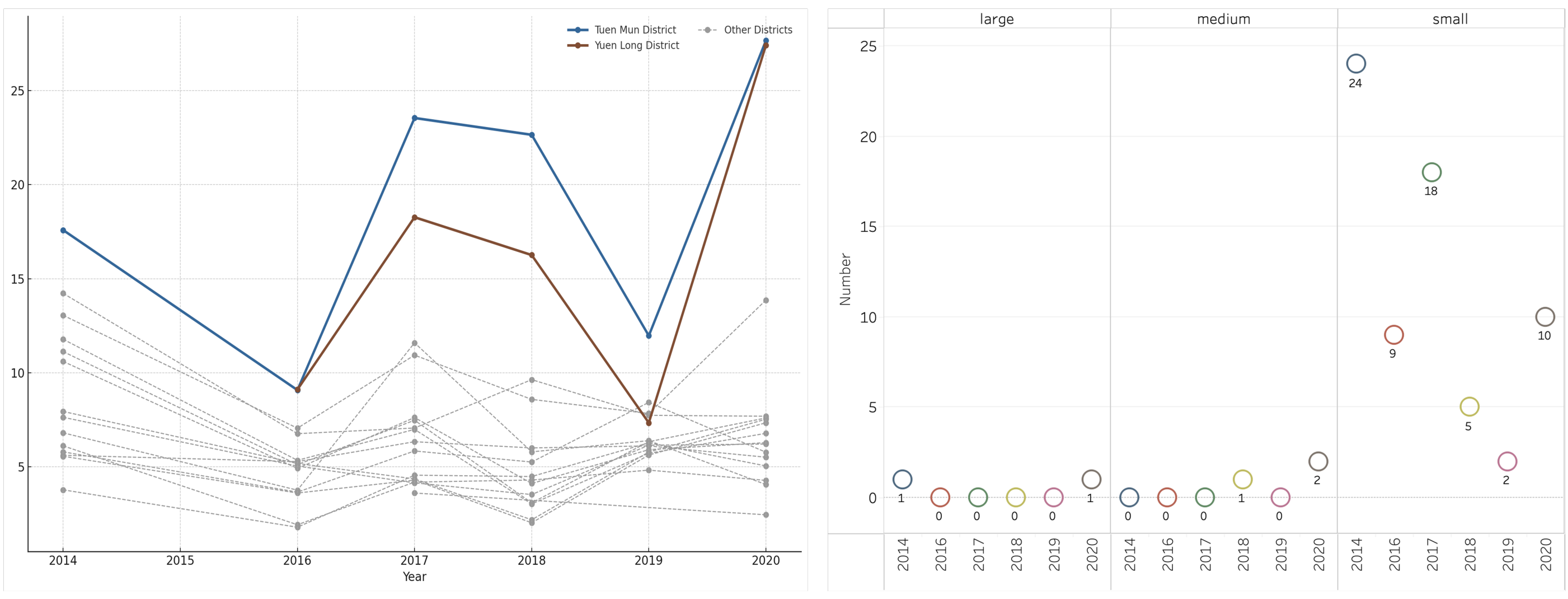

(c) Distribution Map of Heatpoint Clusters in 2018 and 2020

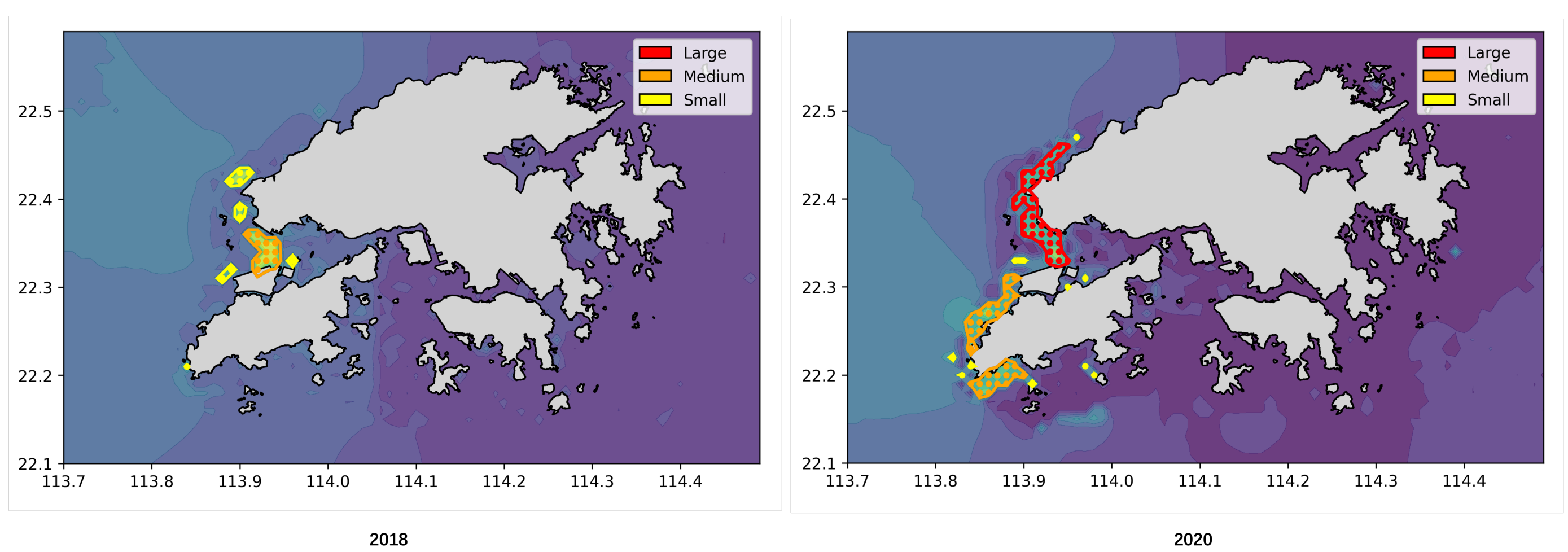

Fig. 7 Analysis of Changes in Average Coastal Water Quality and Heatpoint Cluster Distribution Map in Hong Kong (2014-2020)

Figure 7(a) reveals variations in average water quality across different districts in Hong Kong from 2014 to 2020, excluding 2015 due to missing data. Notably, the Tuen Mun District and Yuen Long District consistently displayed poorer average water quality compared to other districts. Furthermore, the analysis indicates that 2017, 2018, and 2020 were marked by particularly poor average water quality across the region. This temporal concentration hints at potential external influences, such as regional development activities, climatic events, or pollution incidents, warranting further investigation to identify the underlying causes.

Beyond examining average water quality change, we further analyze heatpoint clusters, focusing on areas with concentrated poor water quality. Figure 7(b) highlights the large number of small-sized clusters in 2014 and 2017, indicating concentrated zones of water quality deterioration during these years. In contrast, medium and large size clusters were relatively rare, suggesting that widespread water quality issues were not a predominant concern during most of the analyzed period. The predominance of small clusters in specific years may point to localized pollution sources or discrete environmental events affecting particular water bodies, thereby emphasizing the importance of targeted intervention strategies in affected areas.

Given the emergence of medium and large-sized heatpoint clusters in 2018 and 2020, we queried the MMUEChange agent to generate distribution maps to visualize these areas. As shown in Figure 7(c), these clusters exhibit a noticeable spatial expansion from the Yuen Long District towards the Islands District. This trend may indicate the progressive spread of water quality issues, such as tidal currents or riverine inflows. The spatial diffusion of these poor-quality clusters highlights a potential inter-district connection in water quality patterns, raising concerns about the effectiveness of local water management practices and the need for coordinated efforts to address cross-district pollution sources.

Overall, these insights underscore the value of using the MMUEChange agent to identify, analyze, and interpret complex change trends in water quality data, providing a foundation for decision-making and strategic planning.

### 4.4. Case study 3: Open dumpsite Change and Influential Factor Analysis in Shenzhen

#### 4.4.1. Data and Workflow Introduction

The analysis of open dumpsite changes and the identification of influential factors that drive these changes are crucial for effective urban waste management and sustainable city planning. Open dumpsites, especially in rapidly urbanizing areas like Shenzhen, present significant environmental and public health challenges. Understanding how these dumpsites change and identifying the underlying factors contributing to their growth or decline can help policymakers develop more targeted interventions to mitigate environmental degradation. In this study, we focus on the Baoan District in Shenzhen. As illustrated in Figure 8, remote sensing images are used to detect open dumpsites [70, 71]. The detection results contain domestic waste and construction waste, and only results with a confidence score higher than 0.5 are retained. Then the results are stored in a CSV file for further analysis. In addition to the waste data, the case study also incorporates population data [72], points of interest (POI) data [73], and nighttime light index data [74] collected across different years for Shenzhen, stored in CSV files. These datasets are essential for exploring the broader socioeconomic and urban development factors that may influence changes in open dumpsites.

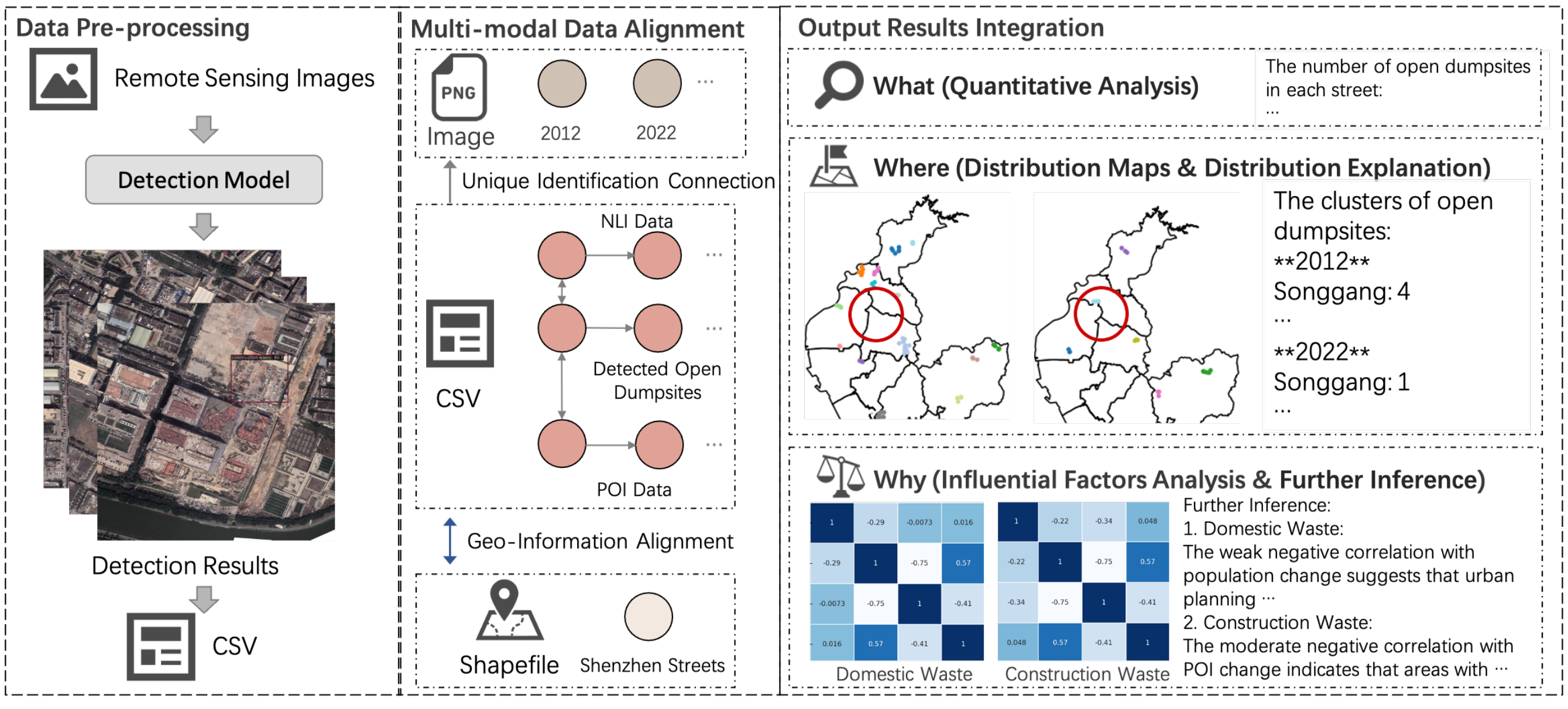

Fig. 8 The overall workflow in case study 3.

The question dataset for this case study is structured across three change analysis levels: 'what', 'where', and 'why'. Specifically, questions in the 'where' level often involve the detection and analysis of dumpsite clusters, generated using the DBSCAN clustering algorithm [75-81]. These clusters represent concentrated areas of waste accumulation, which are visualized in spatial distribution maps. The agent should not only generate these maps but also provide descriptive insights into the spatial patterns observed, such as identifying the changes in regions where dumpsites are densely concentrated. It is important to monitor the changes in dumpsite clusters over time for identifying regions that may require targeted intervention.

At the most complex level, the 'why' level, the agent is required to perform correlation analysis between the various datasets, such as population change, POI number, and nighttime light intensity, and the changes in the number of open dumpsites. This analysis helps to identify potential relationships between urban development factors and the formation or disappearance of open dumpsites. Based on these correlations, the model should also provide further inference, organizing conclusive results based on historical trends.

### 4.4.2. Experimental Performance

The experiment compares the performance of the MMUEChange agent with these baseline models: Standalone Agent, No-alignment Agent, Data-only Agent, and Single-modality Agent. As demonstrated in Table 6, the Standalone Agent consistently generated fabricated or overly generic responses due to the absence of both datasets and tools, while the No-alignment Agent, despite having full data and tools, failed to maintain cross-modal coherence, frequently mismatching identifiers and producing fragmented results. The Data-only Agent performed poorly due to the issues of hallucination, especially when handling and aligning complex, multi-modal data such as shapefiles, resulting in inaccurate answers across all question levels. The Single-modality Agent, while capable of addressing 'what' level questions, such as quantifying waste changes over time, was unable to process the distribution maps generation and explanation tasks required at the 'where' level. Furthermore, its inability to utilize relevant analytical methods meant it could not handle complex tasks such as correlation analysis and inference tasks needed for the 'why' level.

Table 6 Comparison of performance of our MMUEChange Agent and other baseline models in case study 3.

| Change Analysis Level | What | | | | Where | | | | Why | | Accuracy (Correct / Total) |
|---|---|---|---|---|---|---|---|---|---|---|---|
| Subtype | Quantitative | | | | Distribution Maps | | Distribution Explanation | | Influential Factors Analysis | | / |
| Models \ Questions | Q1 | Q2 | Q3 | Q4 | Q5 | Q6 | Q7 | Q8 | Q9 | Q10 | / |
| Standalone Agent | × | × | × | × | × | × | × | × | × | × | 0/10 |
| No-alignment Agent | × | × | × | × | × | × | × | × | × | × | 0/10 |
| Data-only Agent | × | × | × | × | × | × | × | × | × | × | 0/10 |
| Single-modality Agent | ✓ | ✓ | ✓ | ✓ | × | × | × | × | × | × | 4/10 |
| MMUEChange Agent | ✓ | ✓ | ✓ | ✓ | ✓ | ✓ | ✓ | ✓ | ✓ | ✓ | 10/10 |

In contrast, the MMUEChange agent is capable of accurately answering questions across all three levels of analysis. The agent is able to integrate the CSV files and shapefiles to provide a comprehensive analysis of open dumpsite changes. At the 'where' level, the agent successfully generates spatial maps of dumpsite clusters and provided detailed interpretations of the changes observed. At the 'why' level, the agent performs influential factor analysis, linking changes in dumpsites with influential factors such as population growth, urban social development patterns (POIs), and economic development (increased nighttime activity).

#### 4.4.3. Key Findings Discussion

Leveraging the question dataset, we employed the MMUEChange agent to obtain information across three analytical levels. Based on these findings, we derived several key insights into the trends and dynamics of open dumpsites in Shenzhen Baoan District between 2012 and 2022, revealing changes in waste management practices and the factors influencing these changes.

Figure 9 (a) presents the total number of open dumpsites detected in Baoan District in 2012 and 2022, alongside a breakdown of waste types into domestic and construction waste. The results indicate a significant decline of 36.42% in the overall number of open dumpsites from 2012 to 2022, reflecting notable improvements in waste management over the decade. Additionally, the proportion of construction waste decreased from 33% in 2012 to 20% in 2022. This reduction may be attributed to the rapid urban development in this district. The street-level analysis, as illustrated in Figure 9 (b), reveals that most streets experienced slight declines in the number of open dumpsites over this period. However, a few exceptions, such as Fuyong and Yanluo Streets, recorded minor increases. These localized variations might stem from demographic shifts or differing levels of economic activity. The overall pattern indicates that while broad improvements have been made, continuous monitoring and targeted interventions in areas with persistent issues are still necessary.

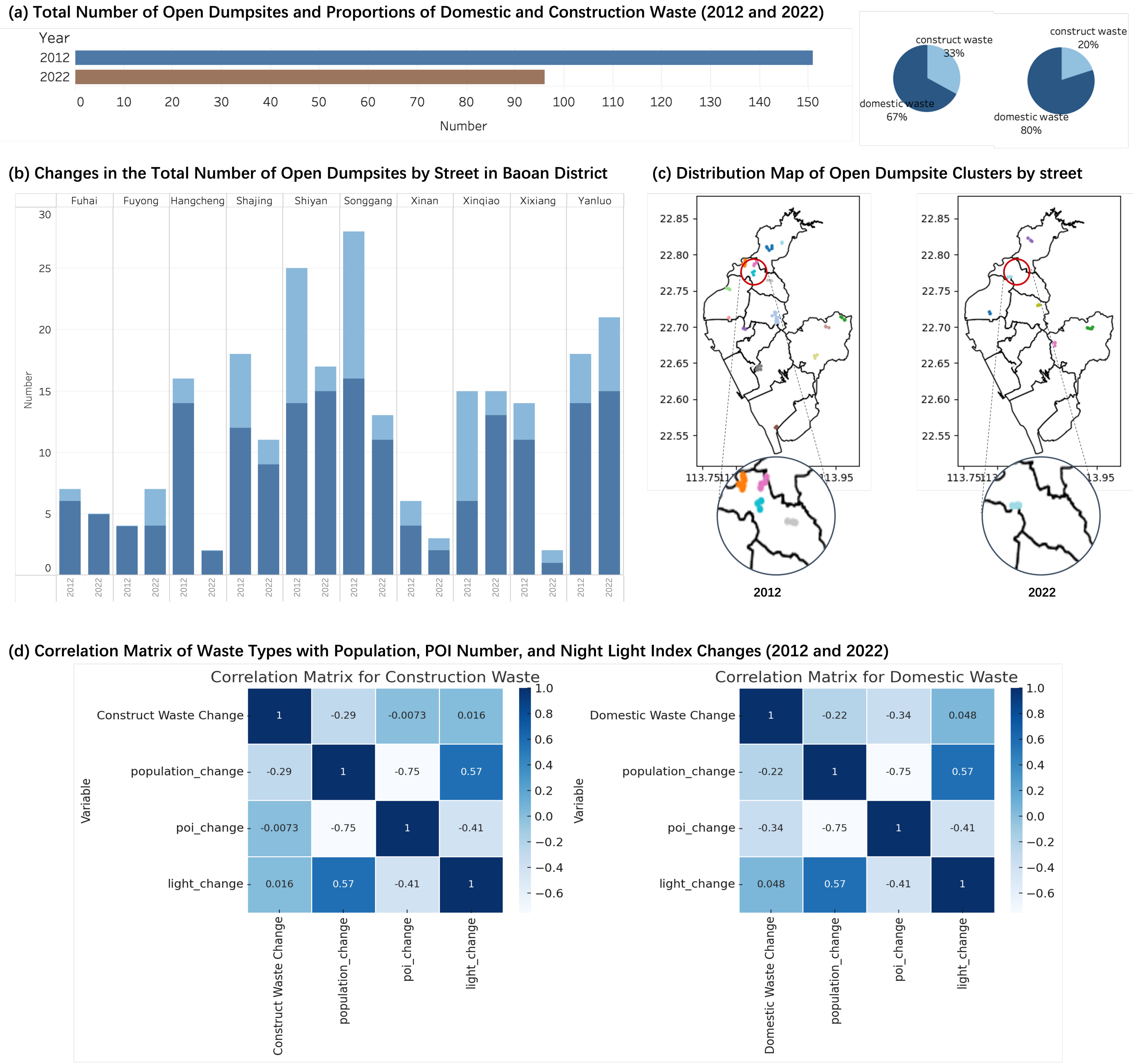

Fig. 9 Analysis of Open Dumpsite Changes in Baoan District, Shenzhen (2012 and 2022)

We also obtained information on the spatial clusters of open dumpsites in Baoan District. Figure 9 (c) illustrates that in 2012, open dumpsites were heavily concentrated in Songgang Street. By 2022, substantial improvements were observed, with a notable reduction in the number of clusters. This positive shift could be attributed to targeted waste management initiatives, increased public awareness, or enhanced regulatory frameworks in the district. The reduction in clustering suggests that waste disposal practices have become more decentralized and controlled, pointing to the effectiveness of waste management strategies in mitigating concentrated dumping issues.

Finally, in Figure 9 (d), the results are presented through an influential factor analysis performed by the MMUEChange agent. The correlation matrices for construction and domestic waste changes reveal several significant insights. For construction waste, there is a weak negative correlation with population change, indicating that increases in population may have a minor reducing effect on construction waste issues, possibly due to enhanced waste infrastructure accompanying urban growth. The lack of strong correlations with POI change and night light index suggests that construction waste changes are less directly influenced by shifts in business density or nighttime economic activities. In contrast, for domestic waste, the correlation with POI change is moderately negative, indicating that areas with increased commercial activity might experience challenges in managing domestic waste effectively. The weak positive correlation between domestic waste change and night light index suggests that increased nighttime activities may be marginally associated with higher levels of domestic waste generation, possibly due to the expansion of retail services.

Overall, these insights reveal the challenges in waste management in the Baoan District. They point to a complex interaction between urbanization, economic activities, and waste disposal practices, highlighting the importance of comprehensive and adaptive strategies in addressing both localized and district-wide waste management issues. Moreover, this case study highlights the strengths of the MMUEChange agent in dealing with diverse and complex data sources, as well as its ability to provide comprehensive responses with multi-level change analysis questions using diverse analytical methods.

### 4.5. Ablation study

To better understand the contribution of individual components within the MMUEChange framework, we conduct an ablation study comparing the full system against several systematically degraded variants. Each ablated agent disables or removes specific modules while keeping all other settings constant. The evaluation is performed on the case study question sets described earlier, with performance measured at three change analysis levels. Table 7 reports the comparative results of the MMUEChange agent and four ablated variants, evaluated across these three levels. The performance is measured in terms of the number of correctly answered questions at each level. Compared to the best-performing baseline agent, the MMUEChange agent achieves a 46.7% improvement in overall task completion.

Table 7 Ablation study Comparing MMUEChange and Variants across Analysis Levels

| Agents | Without | "What" Level | "Where" Level | "Why" Level | Overall |
|---|---|---|---|---|---|
| Standalone Agent | Data & Toolkit | 0 | 0 | 0 | 0 |
| No-alignment Agent | Modality Controller | 0 | 0 | 0 | 0 |
| Data-only Agent | Toolkit | 4/20 | 0/8 | 0/2 | 4/30 |
| Single-modality Agent | Part of Toolkit | 16/20 | 0/8 | 0/2 | 16/30 |
| MMUEChange Agent | / | 20/20 | 8/8 | 2/2 | 30/30 |

Table 7 summarizes the results across five agent configurations:

- Standalone Agent: Operates without any external data or tools. Its outputs rely solely on the LLM's internal knowledge. This setting fails entirely across all reasoning levels, exhibiting severe hallucinations and an inability to ground responses in context.
- No-alignment Agent: Includes the full set of multi-modal tools but disables the Modality Controller responsible for aligning data representations across processing stages and modalities. Without proper data linking and coordination, the agent often fails to retrieve relevant information, resulting in inaccurate or incomplete answers, especially on tasks requiring coherent integration.
- Data-only Agent: Provides access to raw input data but disables external tool usage. The agent can handle a limited subset of descriptive queries by extracting superficial patterns from text, but fails on more complex spatial or inferential tasks due to a lack of structured operations.
- Single-modality Agent: Equipped with tools tailored to a single modality (e.g., CSV), but unable to process or integrate visual or geospatial data. It performs well on "What"-level questions but lacks the capability to handle spatial mapping or multi-modal synthesis.
- MMUEChange Agent: The complete system includes all components: multi-modal data, the modular toolkit, and the Modality Controller. The results demonstrate the necessity of an accurate toolkit for domain-specific information and explicit data alignment.

Overall, the ablation study demonstrates that MMUEChange's performance emerges from the orchestration of the modular toolkit, modality-specific processing, and alignment strategies.

### 4.6. Discussion

#### 4.6.1. Hallucination Issues in the Data-only Agent: Types and Causes

In our experiments, there are significant hallucination issues with the Data-only Agent, which severely impacted its ability to handle complex, multi-modal data. Here, we will discuss several types of hallucination observed and analyze their causes.

The first type of hallucination arises when the Data-only Agent attempts to process multiple CSV files that are linked by a foreign key. In such cases, the Data-only Agent lacks data schema information and does not have a deep understanding of the relationships between the data. Consequently, when generating Python code to handle these CSV files, it often misidentifies the correct foreign key, resulting in errors during data joins, particularly in complex scenarios where multi-CSV table querying is required, such as in case study 1, where multiple CSVs need to be linked for accurate qualitative analysis.

Another issue arises when the Data-only Agent is used to process shapefile data. The model generates Python code that requires importing third-party libraries for geospatial data handling, such as geopandas or shapely. However, due to package version conflicts, the generated code often calls outdated methods that no longer exist in the updated versions of these libraries. For example, a class or function used in the generated code might have been deprecated or changed in newer versions of the package. This mismatch between the current environment and the generated code leads to errors in Python code, such as in the 'what' questions in case studies 2 and 3.

Additionally, the Data-only Agent is constrained by the file size limits from its file-uploading function. For example, in case study 1, the LiDAR data files are too large for the Data-only Agent to be uploaded and process. This limitation restricts its ability to perform complex analyses on large datasets, such as high-resolution remote sensing data, and leads to hallucination when the model attempts to proceed without access to the full dataset.

In summary, using the Data-only Agent for data analysis introduces several significant problems that lead to hallucination, including misalignment of foreign key relations in structured data, package version conflicts in code generation, and file size limitations that prevent comprehensive data processing. Due to these limitations, the Data-only Agent becomes unreliable for complex tasks like urban environment change analysis, where multi-modal data and precise data handling are essential.

#### 4.6.2. Limitations of LangChain Default Agents

In our experiments, structured data are frequently utilized, particularly CSV files, which can be stored and accessed in various formats such as databases, pandas dataframes, or directly as CSV files. LangChain default agents, including the SQL Agent, Pandas Dataframe Agent, and CSV Agent, provide an direct and straightforward method to construct agents to handle these data through pre-defined default processing tools. Although these agents offer a simple way to interact with structured data, they are inherently limited by their fixed data handling workflows and general processing tools.

The rigid workflows of these agents present significant challenges in prompt construction. For instance, when agents attempt to generate prompts for backend LLMs with table data information, they often copy the contents directly into the prompt. This can lead to issues where the prompt exceeds the token limit, resulting in errors. Additionally, a major limitation of these default agents is their inability to track the data they process, making it difficult to handle complex data transformations or multi-step processing tasks. The tools employed by these agents do not maintain a record of changes made to the data throughout the processing pipeline, leading to potential issues with data integrity. In some cases, this lack of data lineage tracking can cause data contamination or loss. For example, if an agent unintentionally modifies a CSV file during processing, it could corrupt the original dataset, leading to inaccurate or incomplete analysis. This becomes particularly problematic in scenarios that require extensive, multi-step data processing workflows.

Furthermore, the tools used by these agents are generally designed for broad applicability rather than tailored to specific use cases. Although they can perform basic tasks, such as querying databases or reading CSV files, they often struggle with more complex multi-modal data processing requirements.

In conclusion, although LangChain default agents provide a straightforward means to construct agents for interacting with structured data, their generalized nature and fixed workflows limit their effectiveness in handling more complex data processing tasks, especially in contexts like urban environment change analysis. Additionally, these agents are prone to data contamination and struggle with multi-modal data integration, making them unreliable for complex applications.

#### 4.6.3. Limitations of Single-Modality Agents

Single-modality agents, which are designed to handle a single type of data, such as CSV files, can perform well in simple data analysis tasks but fall short when applied to complex urban environment change problems.

These agents can efficiently perform tasks like calculating summary statistics or identifying trends in tabular data. However, urban environmental analysis frequently requires insights that go beyond what a single data modality can offer.

For example, when conducting spatial distribution analysis, such as identifying heatpoint clusters of poor water quality in case study 2, distribution maps are essential for understanding spatial patterns. These maps provide a direct reference that cannot be conveyed through tabular data alone. A single-modality agent, which is limited to processing CSV files, cannot generate these visual outputs or interpret spatial patterns effectively.

In conclusion, while single-modality agents can handle basic CSV data analysis, they are insufficient for addressing complex urban environment change problems that require multi-modal data processing and spatial analysis. The ability to align multi-modal data and choose the appropriate analytical methods for each modality is essential for comprehensive urban research, and single-modality agents lack the flexibility and capability to meet these demands.

#### 4.6.4. Computational Cost, Scalability, and Optimization Strategies for Real-World Applications

The deployment of the MMUEChange agent inevitably involves trade-offs between analytical depth and computational performance. In our current implementation, the LLM backend operates via cloud-based API services, which enables access to advanced reasoning capabilities without the need for maintaining local model infrastructure. More broadly, two deployment paradigms can be considered: local inference, which offers lower network dependency and stronger privacy but requires significant computing resources for hosting large models, and API-based inference, which reduces local hardware demands but entails higher external dependency and variable latency. The performance estimates reported below are therefore grounded in the API-based backend configuration of our MMUEChange agent.

End-to-end latency arises from two primary sources: the toolkit modules, which process heterogeneous data modalities, and the LLM backend, which coordinates tool calls, manages reasoning chains, and synthesizes results. Different modalities incur distinct computational burdens. Structured tabular data are lightweight and efficient to process, while spatial vector data (e.g., shapefiles) require additional operations such as projection or overlay. High-volume modalities such as LiDAR demand heavier preprocessing and indexing. Meanwhile, image generation and interpretation can be computationally intensive. Analytical modules such as Influential

Factors Analysis are relatively lightweight per call but may increase the overall coordination overhead by triggering additional reasoning steps.

Beyond tool execution, the number of LLM coordination rounds constitutes the dominant non-tool cost. Each round encompasses planning, invoking tools, reflecting on outputs, and composing answers. As task complexity increases from what to where to why levels of analysis, the required rounds grow accordingly. This scaling effect implies that overall response time is shaped not only by the cost of individual tool operations but also by the orchestration burden imposed on the LLM backend.

The estimates in Table 8 illustrate how computational performance varies across the three levels of analysis. At the what level, latency remains low since only lightweight tabular data and limited LiDAR operations are involved. The where level introduces spatial vector data and image-based operations, substantially raising tool-side costs while requiring additional LLM coordination. At the why level, the inclusion of explanatory modules such as Influential Factors Analysis increases both the reasoning depth and the number of LLM rounds, resulting in high end-to-end latency.

Table 8. Estimated computational cost and latency of the MMUEChange agent under varying analysis levels.

| Analysis Level | Typical Data Modalities (in Case Studies) | Typical Tool Calls by Modality | Avg. Tool Call Latency (ms) | Typical LLM API Rounds | Avg. API Time per Call(s) | Estimated Total Latency(s/ query) |
|---|---|---|---|---|---|---|
| **What** | CSV (NYC), LiDAR (small coverage) | CSV 4–8, LiDAR 2 | CSV 50–100; LiDAR 800-1000; Shapefile 500–600; Image 600–800; Factors 50–100 | 6-10 | 1.5–3.0 | ~10.8–32.8 |
| **Where** | CSV + Shapefile + Image Generation | CSV 4–6, Shapefile 2, Image 2 | | 8-10 | 1.5–3.0 | ~14.4–33.4 |
| **Why** | CSV + Shapefile + Image Generation + Influential Factors Analysis | CSV 4–6, Shapefile 2, Image 2, Factors 2 | | 10-12 | 1.5–3.0 | ~17.5–39.6 |

Beyond the computational footprint of individual queries, a central concern for the MMUEChange agent is scalability when applied to city-scale or longitudinal datasets. As the scope of analysis expands, the cumulative demand on both toolkit modules and the LLM backend grows non-linearly. For example, large-scale LiDAR collections require extensive tiling and indexing, while thousands of shapefile overlays or image-generation requests may quickly saturate computational pipelines. Moreover, higher-level why analyses, which inherently involve iterative reasoning and multi-step synthesis, amplify the number of LLM rounds required and thus extend overall latency. These characteristics imply that straightforward scaling from small case studies to full metropolitan deployments is not trivial. Instead, careful system-level design is necessary to ensure responsiveness and throughput remain acceptable as data volume and query complexity increase.

To address these scalability challenges, a set of optimization strategies can be adopted at multiple layers of the system. At the data layer, pre-processing heavy modalities (e.g., tiling LiDAR into manageable spatial units, pre-indexing shapefiles, and caching commonly used CSV summaries) can reduce redundant computation. At the deployment layer, the trade-offs between local inference and API services become central. Local inference offers lower latency and stronger control over data privacy but requires substantial computing infrastructure, while cloud APIs enable access to advanced models with reduced local cost but higher dependency on external providers. A hybrid strategy that combines the strengths of both, using local models for frequent, low-complexity operations and reserving cloud APIs for occasional, high-complexity analyses, can therefore help balance latency, cost efficiency, and accuracy. Together, these measures provide a pathway for translating the MMUEChange agent from research prototypes to scalable, real-world applications in urban change monitoring.

#### 4.6.5. Potential Policy Implications

The three case studies presented in this work illustrate not only the technical versatility of the MMUEChange agent but also its potential value as a decision-support tool for urban policy and planning. By fusing heterogeneous data streams and generating interpretable outputs across different analytical levels, MMUEChange enables a range of evidence-based interventions that extend beyond academic research into practical governance.

In New York, the agent was applied to describe newly constructed urban parks by integrating tabular property data with LiDAR-derived structural information. This capability can inform urban greening strategies, for example, helping municipal authorities identify which neighborhoods have received limited park development or where park facilities remain underutilized. Such insights can guide more equitable allocation of green infrastructure, improve accessibility, and support long-term environmental and public health objectives.

In Hong Kong, the agent was used to monitor coastal water quality change by linking water parameter datasets with district-level boundaries. This integration provides actionable intelligence for environmental regulation and compliance monitoring, enabling agencies to pinpoint emerging pollution hotspots, evaluate the effectiveness of mitigation measures, and communicate findings transparently to stakeholders. Over time, this function could be embedded into real-time coastal management systems, supporting adaptive policy responses to dynamic environmental risks.

In Shenzhen, the agent analyzed the dynamics of open dumpsite change and its influential factors, correlating population shifts, POI density, and nighttime light intensity with the emergence or disappearance of waste sites. This approach can support waste management and urban sanitation policies, such as prioritizing enforcement or infrastructure investments in areas most at risk of illegal dumping. More broadly, it provides a scalable template for linking socio-economic indicators to environmental outcomes, thereby assisting in anticipatory planning and targeted interventions.

The modular and interpretable design of the agent makes it suitable for integration into urban digital twins and decision-support dashboards. By incorporating additional real-time data streams (e.g., traffic sensors, social media reports) and domain-specific analytical modules, the agent could evolve into a flexible backend for policy simulation, scenario planning, and participatory governance. This potential underscores its relevance not only as a research framework but also as a practical enabler of data-driven, transparent, and adaptive urban planning.

### 4.7. Limitation

Despite the demonstrated adaptability and effectiveness of the proposed framework, several limitations remain that warrant discussion.

First, the framework inherently relies on the reasoning capabilities of the underlying LLM. While our design incorporates alignment mechanisms and domain-specific toolkits, the depth and stability of reasoning ultimately depend on the backend model, which may occasionally generate inconsistent or suboptimal plans.

Second, although the modular toolkit was developed to ensure flexibility, its scalability poses practical challenges. As the number and diversity of modalities increase, maintaining a large collection of specialized modules introduces computational overhead and requires continuous updates to remain effective across domains. Third, the framework still requires careful prompt engineering to ensure robust performance. Designing prompts that balance domain specificity with generalization remains non-trivial, and excessive reliance on handcrafted prompting strategies may limit ease of use for non-expert practitioners.

These limitations are natural consequences of applying emerging LLM-based systems. Future research should focus on mitigating them through the integration of more transparent reasoning engines, the design of lightweight yet extensible toolkits, and the development of adaptive prompting strategies that reduce reliance on manual intervention.

## 5. Conclusion

This study proposed a generalized agent framework for multi-modal urban environment change analysis and instantiated it through the MMUEChange agent, which was validated across three case studies representing different analytical levels and urban contexts. The results demonstrated that the agent can uncover strategic patterns in park development in New York, identify spatial clusters of coastal water degradation in Hong Kong, and reveal the socioeconomic drivers behind waste reduction in Shenzhen, thereby showing its capacity to move from descriptive observations to causal reasoning through the hierarchical "what–where–why" paradigm. By systematically aligning user queries, selecting appropriate modalities, and mitigating hallucination through a modular toolkit, the framework consistently outperformed baseline LLMs and delivered interpretable insights. More broadly, these findings illustrate how LLM agents can transform urban study by advancing beyond pixel-level or single-modality analyses to provide integrative, transparent, and decision-oriented knowledge. Such capabilities hold promise for policymakers, offering a pathway to evidence-based strategies for sustainable urban planning, resilient infrastructure management, and adaptive governance in the face of accelerating environment change.

**Declaration of Generative AI and AI-assisted technologies in the writing process**

Statement: During the preparation of this work, the author(s) used ChatGPT-4 in order to improve readability and language. After using this tool/service, the author(s) reviewed and edited the content as needed and take(s) full responsibility for the content of the publication.

## References

[1] Y. Himeur, B. Rimal, A. Tiwary, A. Amira, Using artificial intelligence and data fusion for environmental monitoring: A review and future perspectives, Information Fusion 86–87 (2022) 44–75. https://doi.org/10.1016/j.inffus.2022.06.003.

[2] K. Ezimand, H. Aghighi, D. Ashourloo, A. Shakiba, The analysis of the spatio-temporal changes and prediction of built-up lands and urban heat islands using multi-temporal satellite imagery, Sustainable Cities and Society 103 (2024) 105231. https://doi.org/10.1016/j.scs.2024.105231.

[3] W. Li, Z. Cai, L. Jin, Urban green land use efficiency of resource-based cities in China: Multidimensional measurements, spatial-temporal changes, and driving factors, Sustainable Cities and Society 104 (2024) 105299. https://doi.org/10.1016/j.scs.2024.105299.

[4] H. Wu, F. Wu, Y. Cai, Z. Li, Assessing the spatiotemporal impacts of land use change on ecological environmental quality using a regionalized territorial impact assessment framework, Sustainable Cities and Society 112 (2024) 105623. https://doi.org/10.1016/j.scs.2024.105623.

[5] V. Amini Parsa, E. Salehi, A.R. Yavari, P.M. Van Bodegom, Analyzing temporal changes in urban forest structure and the effect on air quality improvement, Sustainable Cities and Society 48 (2019) 101548. https://doi.org/10.1016/j.scs.2019.101548.

[6] Y. Wang, Y. Zhang, W. Sun, L. Zhu, The impact of new urbanization and industrial structural changes on regional water stress based on water footprints, Sustainable Cities and Society 79 (2022) 103686. https://doi.org/10.1016/j.scs.2022.103686.

[7] V. Obradovic, A. Vulevic, Water Resources Protection and Water Management Framework in Western Balkan Countries in Drina River Basin, ATG 2 (2023) 24–32. https://doi.org/10.56578/atg020103.

[8] R.L. Mahler, Public Perceptions and Evaluations of Drinking Water Quality in Idaho: A 35-Year Survey Analysis, OCS 3 (2024) 147–157. https://doi.org/10.56578/ocs030301.

[9] J. Karmakar, Transformation of Urban Pondscapes in the Kolkata Metropolitan Area: A Case Study of Serampore Municipality, SSRN Journal (2025). https://doi.org/10.2139/ssrn.5136455.

[10] M. Reba, K.C. Seto, A systematic review and assessment of algorithms to detect, characterize, and monitor urban land change, Remote Sensing of Environment 242 (2020) 111739. https://doi.org/10.1016/j.rse.2020.111739.

[11] Z. Xiao, J. Qi, W. Xue, P. Zhong, Few-shot object detection with self-adaptive attention network for remote sensing images, IEEE Journal of Selected Topics in Applied Earth Observations and Remote Sensing 14 (2021) 4854–4865.

[12] Z. Xiao, P. Zhong, Y. Quan, X. Yin, W. Xue, Few-shot object detection with feature attention highlight module in remote sensing images, in: 2020 International Conference on Image, Video Processing and Artificial Intelligence, International Society for Optics and Photonics, 2020: p. 115840Z. https://doi.org/10.1117/12.2577473

[13] OpenAI, J. Achiam, S. Adler, S. Agarwal, L. Ahmad, I. Akkaya, F.L. Aleman, D. Almeida, J. Altenschmidt, S. Altman, S. Anadkat, R. Avila, I. Babuschkin, S. Balaji, V. Balcom, P. Baltescu, H. Bao, M. Bavarian, J. Belgum, I. Bello, J. Berdine, G. Bernadett-Shapiro, C. Berner, L. Bogdonoff, O. Boiko, M. Boyd, A.-L. Brakman, G. Brockman, T. Brooks, M. Brundage, K. Button, T. Cai, R. Campbell, A. Cann, B. Carey, C. Carlson, R. Carmichael, B. Chan, C. Chang, F. Chantzis, D. Chen, S. Chen, R. Chen, J. Chen, M. Chen, B. Chess, C. Cho, C. Chu, H.W. Chung, D. Cummings, J. Currier, Y. Dai, C. Decareaux, T. Degry, N. Deutsch, D. Deville, A. Dhar, D. Dohan, S. Dowling, S. Dunning, A. Ecoffet, A. Eleti, T. Eloundou, D. Farhi, L. Fedus, N. Felix, S.P. Fishman, J. Forte, I. Fulford, L. Gao, E. Georges, C. Gibson, V. Goel, T. Gogineni, G. Goh, R. Gontijo-Lopes, J. Gordon, M. Grafstein, S. Gray, R. Greene, J. Gross, S.S. Gu, Y. Guo, C. Hallacy, J. Han, J. Harris, Y. He, M. Heaton, J. Heidecke, C. Hesse, A. Hickey, W. Hickey, P. Hoeschele, B. Houghton, K. Hsu, S. Hu, X. Hu, J. Huizinga, S. Jain, S. Jain, J. Jang, A. Jiang, R. Jiang, H. Jin, D. Jin, S. Jomoto, B. Jonn, H. Jun, T. Kaftan, Ł. Kaiser, A. Kamali, I. Kanitscheider, N.S. Keskar, T. Khan, L. Kilpatrick, J.W. Kim, C. Kim, Y. Kim, J.H. Kirchner, J. Kiros, M. Knight, D. Kokotajlo, Ł. Kondraciuk, A. Kondrich, A. Konstantinidis, K. Kosic, G. Krueger, V. Kuo, M. Lampe, I. Lan, T. Lee, J. Leike, J. Leung, D. Levy, C.M. Li, R. Lim, M. Lin, S. Lin, M. Litwin, T. Lopez, R. Lowe, P. Lue, A. Makanju, K. Malfacini, S. Manning, T. Markov, Y. Markovski, B. Martin, K. Mayer, A. Mayne, B. McGrew, S.M. McKinney, C. McLeavey, P. McMillan, J. McNeil, D. Medina, A. Mehta, J. Menick, L. Metz, A. Mishchenko, P.

Mishkin, V. Monaco, E. Morikawa, D. Mossing, T. Mu, M. Murati, O. Murk, D. Mély, A. Nair, R. Nakano, R. Nayak, A. Neelakantan, R. Ngo, H. Noh, L. Ouyang, C. O'Keefe, J. Pachocki, A. Paino, J. Palermo, A. Pantuliano, G. Parascandolo, J. Parish, E. Parparita, A. Passos, M. Pavlov, A. Peng, A. Perelman, F. de A.B. Peres, M. Petrov, H.P. de O. Pinto, Michael, Pokorny, M. Pokrass, V.H. Pong, T. Powell, A. Power, B. Power, E. Proehl, R. Puri, A. Radford, J. Rae, A. Ramesh, C. Raymond, F. Real, K. Rimbach, C. Ross, B. Rotsted, H. Roussez, N. Ryder, M. Saltarelli, T. Sanders, S. Santurkar, G. Sastry, H. Schmidt, D. Schnurr, J. Schulman, D. Selsam, K. Sheppard, T. Sherbakov, J. Shieh, S. Shoker, P. Shyam, S. Sidor, E. Sigler, M. Simens, J. Sitkin, K. Slama, I. Sohl, B. Sokolowsky, Y. Song, N. Staudacher, F.P. Such, N. Summers, I. Sutskever, J. Tang, N. Tezak, M.B. Thompson, P. Tillet, A. Tootoonchian, E. Tseng, P. Tuggle, N. Turley, J. Tworek, J.F.C. Uribe, A. Vallone, A. Vijayvergiya, C. Voss, C. Wainwright, J.J. Wang, A. Wang, B. Wang, J. Ward, J. Wei, C.J. Weinmann, A. Welihinda, P. Welinder, J. Weng, L. Weng, M. Wiethoff, D. Willner, C. Winter, S. Wolrich, H. Wong, L. Workman, S. Wu, J. Wu, M. Wu, K. Xiao, T. Xu, S. Yoo, K. Yu, Q. Yuan, W. Zaremba, R. Zellers, C. Zhang, M. Zhang, S. Zhao, T. Zheng, J. Zhuang, W. Zhuk, B. Zoph, GPT-4 Technical Report, (2024). http://arxiv.org/abs/2303.08774 (accessed October 18, 2024).

[14] E.J. Parelius, A Review of Deep-Learning Methods for Change Detection in Multispectral Remote Sensing Images, Remote Sensing 15 (2023). https://doi.org/10.3390/rs15082092.

[15] D. Lu, P. Mausel, E. Brondizio, E. Moran, Change detection techniques, International Journal of Remote Sensing 25 (2004) 2365–2401. https://doi.org/10.1080/0143116031000139863

[16] J. Zhao, Y. Chang, J. Yang, Y. Niu, Z. Lu, P. Li, A Novel Change Detection Method Based on Statistical Distribution Characteristics Using Multi-Temporal PolSAR Data, Sensors 20 (2020) 1508. https://doi.org/10.3390/s20051508.

[17] T. Bai, L. Wang, D. Yin, K. Sun, Y. Chen, W. Li, D. Li, Deep learning for change detection in remote sensing: a review, Geo-Spatial Information Science 26 (2023) 262–288. https://doi.org/10.1080/10095020.2022.2085633.

[18] J. Liu, M. Gong, J. Zhao, H. Li, L. Jiao, Difference representation learning using stacked restricted Boltzmann machines for change detection in SAR images, Soft Comput 20 (2016) 4645–4657. https://doi.org/10.1007/s00500-014-1460-0.

[19] R. Xiao, R. Cui, M. Lin, L. Chen, Y. Ni, X. Lin, SOMDNCD: Image Change Detection Based on Self-Organizing Maps and Deep Neural Networks, IEEE Access 6 (2018) 35915–35925. https://doi.org/10.1109/ACCESS.2018.2849110.

[20] P. Zhang, M. Gong, L. Su, J. Liu, Z. Li, Change detection based on deep feature representation and mapping transformation for multi-spatial-resolution remote sensing images, ISPRS Journal of Photogrammetry and Remote Sensing 116 (2016) 24–41. https://doi.org/10.1016/j.isprsjprs.2016.02.013.

[21] T. Zhan, M. Gong, J. Liu, P. Zhang, Iterative feature mapping network for detecting multiple changes in multi-source remote sensing images, ISPRS Journal of Photogrammetry and Remote Sensing 146 (2018) 38–51. https://doi.org/10.1016/j.isprsjprs.2018.09.002.

[22] M. Gong, L. Pan, T. Song, G. Zhang, eds., Bio-inspired Computing – Theories and Applications: 11th International Conference, BIC-TA 2016, Xi'an, China, October 28-30, 2016, Revised Selected Papers, Part II, Springer Singapore, Singapore, 2016. https://doi.org/10.1007/978-981-10-3614-9.

[23] Y. You, J. Cao, W. Zhou, A Survey of Change Detection Methods Based on Remote Sensing Images for Multi-Source and Multi-Objective Scenarios, Remote Sensing 12 (2020) 2460. https://doi.org/10.3390/rs12152460.

[24] J. Liu, M. Gong, K. Qin, P. Zhang, A Deep Convolutional Coupling Network for Change Detection Based on Heterogeneous Optical and Radar Images, IEEE Trans. Neural Netw. Learning Syst. 29 (2018) 545–559. https://doi.org/10.1109/TNNLS.2016.2636227.

[25] A.M. Lal, S.M. Anouncia, Modernizing the multi-temporal multispectral remotely sensed image change detection for global maxima through binary particle swarm optimization, Journal of King Saud University - Computer and Information Sciences 34 (2022) 95–103. https://doi.org/10.1016/j.jksuci.2018.10.010.

[26] A. Dubey, A. Jauhri, A. Pandey, A. Kadian, A. Al-Dahle, A. Letman, A. Mathur, A. Schelten, A. Yang, A. Fan, A. Goyal, A. Hartshorn, A. Yang, A. Mitra, A. Sravankumar, A. Korenev, A. Hinsvark, A. Rao, A. Zhang, A. Rodriguez, A. Gregerson, A. Spataru, B. Roziere, B. Biron, B. Tang, B. Chern, C.

Caucheteux, C. Nayak, C. Bi, C. Marra, C. McConnell, C. Keller, C. Touret, C. Wu, C. Wong, C.C. Ferrer, C. Nikolaidis, D. Allonsius, D. Song, D. Pintz, D. Livshits, D. Esiobu, D. Choudhary, D. Mahajan, D. Garcia-Olano, D. Perino, D. Hupkes, E. Lakomkin, E. AlBadawy, E. Lobanova, E. Dinan, E.M. Smith, F. Radenovic, F. Zhang, G. Synnaeve, G. Lee, G.L. Anderson, G. Nail, G. Mialon, G. Pang, G. Cucurell, H. Nguyen, H. Korevaar, H. Xu, H. Touvron, I. Zarov, I.A. Ibarra, I. Kloumann, I. Misra, I. Evtimov, J. Copet, J. Lee, J. Geffert, J. Vranes, J. Park, J. Mahadeokar, J. Shah, J. van der Linde, J. Billock, J. Hong, J. Lee, J. Fu, J. Chi, J. Huang, J. Liu, J. Wang, J. Yu, J. Bitton, J. Spisak, J. Park, J. Rocca, J. Johnstun, J. Saxe, J. Jia, K.V. Alwala, K. Upasani, K. Plawiak, K. Li, K. Heafield, K. Stone, K. El-Arini, K. Iyer, K. Malik, K. Chiu, K. Bhalla, L. Rantala-Yeary, L. van der Maaten, L. Chen, L. Tan, L. Jenkins, L. Martin, L. Madaan, L. Malo, L. Blecher, L. Landzaat, L. de Oliveira, M. Muzzi, M. Pasupuleti, M. Singh, M. Paluri, M. Kardas, M. Oldham, M. Rita, M. Pavlova, M. Kambadur, M. Lewis, M. Si, M.K. Singh, M. Hassan, N. Goyal, N. Torabi, N. Bashlykov, N. Bogoychev, N. Chatterji, O. Duchenne, O. Çelebi, P. Alrassy, P. Zhang, P. Li, P. Vasic, P. Weng, P. Bhargava, P. Dubal, P. Krishnan, P.S. Koura, P. Xu, Q. He, Q. Dong, R. Srinivasan, R. Ganapathy, R. Calderer, R.S. Cabral, R. Stojnic, R. Raileanu, R. Girdhar, R. Patel, R. Sauvestre, R. Polidoro, R. Sumbaly, R. Taylor, R. Silva, R. Hou, R. Wang, S. Hosseini, S. Chennabasappa, S. Singh, S. Bell, S.S. Kim, S. Edunov, S. Nie, S. Narang, S. Raparthy, S. Shen, S. Wan, S. Bhosale, S. Zhang, S. Vandenhende, S. Batra, S. Whitman, S. Sootla, S. Collot, S. Gururangan, S. Borodinsky, T. Herman, T. Fowler, T. Sheasha, T. Georgiou, T. Scialom, T. Speckbacher, T. Mihaylov, T. Xiao, U. Karn, V. Goswami, V. Gupta, V. Ramanathan, V. Kerkez, V. Gonguet, V. Do, V. Vogeti, V. Petrovic, W. Chu, W. Xiong, W. Fu, W. Meers, X. Martinet, X. Wang, X.E. Tan, X. Xie, X. Jia, X. Wang, Y. Goldschlag, Y. Gaur, Y. Babaei, Y. Wen, Y. Song, Y. Zhang, Y. Li, Y. Mao, Z.D. Coudert, Z. Yan, Z. Chen, Z. Papakipos, A. Singh, A. Grattafiori, A. Jain, A. Kelsey, A. Shajnfeld, A. Gangidi, A. Victoria, A. Goldstand, A. Menon, A. Sharma, A. Boesenberg, A. Vaughan, A. Baevski, A. Feinstein, A. Kallet, A. Sangani, A. Yunus, A. Lupu, A. Alvarado, A. Caples, A. Gu, A. Ho, A. Poulton, A. Ryan, A. Ramchandani, A. Franco, A. Saraf, A. Chowdhury, A. Gabriel, A. Bharambe, A. Eisenman, A. Yazdan, B. James, B. Maurer, B. Leonhardi, B. Huang, B. Loyd, B.D. Paola, B. Paranjape, B. Liu, B. Wu, B. Ni, B. Hancock, B. Wasti, B. Spence, B. Stojkovic, B. Gamido, B. Montalvo, C. Parker, C. Burton, C. Mejia, C. Wang, C. Kim, C. Zhou, C. Hu, C.-H. Chu, C. Cai, C. Tindal, C. Feichtenhofer, D. Civin, D. Beaty, D. Kreymer, D. Li, D. Wyatt, D. Adkins, D. Xu, D. Testuggine, D. David, D. Parikh, D. Liskovich, D. Foss, D. Wang, D. Le, D. Holland, E. Dowling, E. Jamil, E. Montgomery, E. Presani, E. Hahn, E. Wood, E. Brinkman, E. Arcaute, E. Dunbar, E. Smothers, F. Sun, F. Kreuk, F. Tian, F. Ozgenel, F. Caggioni, F. Guzmán, F. Kanayet, F. Seide, G.M. Florez, G. Schwarz, G. Badeer, G. Swee, G. Halpern, G. Thattai, G. Herman, G. Sizov, Guangyi, Zhang, G. Lakshminarayanan, H. Shojanazeri, H. Zou, H. Wang, H. Zha, H. Habeeb, H. Rudolph, H. Suk, H. Aspegren, H. Goldman, I. Damlaj, I. Molybog, I. Tufanov, I.-E. Veliche, I. Gat, J. Weissman, J. Geboski, J. Kohli, J. Asher, J.-B. Gaya, J. Marcus, J. Tang, J. Chan, J. Zhen, J. Reizenstein, J. Teboul, J. Zhong, J. Jin, J. Yang, J. Cummings, J. Carvill, J. Shepard, J. McPhie, J. Torres, J. Ginsburg, J. Wang, K. Wu, K.H. U, K. Saxena, K. Prasad, K. Khandelwal, K. Zand, K. Matosich, K. Veeraraghavan, K. Michelena, K. Li, K. Huang, K. Chawla, K. Lakhotia, K. Huang, L. Chen, L. Garg, L. A, L. Silva, L. Bell, L. Zhang, L. Guo, L. Yu, L. Moshkovich, L. Wehrstedt, M. Khabsa, M. Avalani, M. Bhatt, M. Tsimpoukelli, M. Mankus, M. Hasson, M. Lennie, M. Reso, M. Groshev, M. Naumov, M. Lathi, M. Keneally, M.L. Seltzer, M. Valko, M. Restrepo, M. Patel, M. Vyatskov, M. Samvelyan, M. Clark, M. Macey, M. Wang, M.J. Hermoso, M. Metanat, M. Rastegari, M. Bansal, N. Santhanam, N. Parks, N. White, N. Bawa, N. Singhal, N. Egebo, N. Usunier, N.P. Laptev, N. Dong, N. Zhang, N. Cheng, O. Chernoguz, O. Hart, O. Salpekar, O. Kalinli, P. Kent, P. Parekh, P. Saab, P. Balaji, P. Rittner, P. Bontrager, P. Roux, P. Dollar, P. Zvyagina, P. Ratanchandani, P. Yuvraj, Q. Liang, R. Alao, R. Rodriguez, R. Ayub, R. Murthy, R. Nayani, R. Mitra, R. Li, R. Hogan, R. Battey, R. Wang, R. Maheswari, R. Howes, R. Rinott, S.J. Bondu, S. Datta, S. Chugh, S. Hunt, S. Dhillon, S. Sidorov, S. Pan, S. Verma, S. Yamamoto, S. Ramaswamy, S. Lindsay, S. Lindsay, S. Feng, S. Lin, S.C. Zha, S. Shankar, S. Zhang, S. Zhang, S. Wang, S. Agarwal, S. Sajuyigbe, S. Chintala, S. Max, S. Chen, S. Kehoe, S. Satterfield, S. Govindaprasad, S. Gupta, S. Cho, S. Virk, S. Subramanian, S. Choudhury, S. Goldman, T. Remez, T. Glaser, T. Best, T. Kohler, T. Robinson, T. Li, T. Zhang, T. Matthews, T. Chou, T. Shaked, V. Vontimitta, V. Ajayi, V. Montanez, V. Mohan, V.S. Kumar, V. Mangla, V. Albiero, V. Ionescu, V.

Poenaru, V.T. Mihailescu, V. Ivanov, W. Li, W. Wang, W. Jiang, W. Bouaziz, W. Constable, X. Tang, X. Wang, X. Wu, X. Wang, X. Xia, X. Wu, X. Gao, Y. Chen, Y. Hu, Y. Jia, Y. Qi, Y. Li, Y. Zhang, Y. Zhang, Y. Adi, Y. Nam, Yu, Wang, Y. Hao, Y. Qian, Y. He, Z. Rait, Z. DeVito, Z. Rosnbrick, Z. Wen, Z. Yang, Z. Zhao, The Llama 3 Herd of Models, (2024). http://arxiv.org/abs/2407.21783 (accessed October 18, 2024).

[27] S. Schulhoff, M. Ilie, N. Balepur, K. Kahadze, A. Liu, C. Si, Y. Li, A. Gupta, H. Han, S. Schulhoff, P.S. Dulepet, S. Vidyadhara, D. Ki, S. Agrawal, C. Pham, G. Kroiz, F. Li, H. Tao, A. Srivastava, H. Da Costa, S. Gupta, M.L. Rogers, I. Goncearenco, G. Sarli, I. Galynker, D. Peskoff, M. Carpuat, J. White, S. Anadkat, A. Hoyle, P. Resnik, The Prompt Report: A Systematic Survey of Prompting Techniques, (2024). http://arxiv.org/abs/2406.06608.

[28] X. Wang, J. Wei, D. Schuurmans, Q. Le, E. Chi, S. Narang, A. Chowdhery, D. Zhou, Self-Consistency Improves Chain of Thought Reasoning in Language Models, (2022) 1–24.

[29] Q. Dong, L. Li, D. Dai, C. Zheng, J. Ma, R. Li, H. Xia, J. Xu, Z. Wu, T. Liu, B. Chang, X. Sun, L. Li, Z. Sui, A Survey on In-context Learning, (2024). http://arxiv.org/abs/2301.00234 (accessed October 21, 2024).

[30] O. Rubin, J. Herzig, J. Berant, Learning To Retrieve Prompts for In-Context Learning, NAACL 2022 - 2022 Conference of the North American Chapter of the Association for Computational Linguistics: Human Language Technologies, Proceedings of the Conference (2022) 2655–2671. https://doi.org/10.18653/v1/2022.naacl-main.191.

[31] J. Wei, X. Wang, D. Schuurmans, M. Bosma, B. Ichter, F. Xia, E. Chi, Q. Le, D. Zhou, Chain-of-Thought Prompting Elicits Reasoning in Large Language Models, (2022) 1–14. http://arxiv.org/abs/2203.11171.

[32] T. Kojima, S.S. Gu, M. Reid, Y. Matsuo, Y. Iwasawa, Large Language Models are Zero-Shot Reasoners, (2022). http://arxiv.org/abs/2205.11916.

[33] M. Besta, N. Blach, A. Kubicek, R. Gerstenberger, L. Gianinazzi, J. Gajda, T. Lehmann, M. Podstawski, H. Niewiadomski, P. Nyczyk, T. Hoefler, Graph of Thoughts: Solving Elaborate Problems with Large Language Models, (2023). http://arxiv.org/abs/2308.09687.

[34] J. Long, Large Language Model Guided Tree-of-Thought, (2023). http://arxiv.org/abs/2305.08291.

[35] S. Yao, D. Yu, J. Zhao, I. Shafran, T.L. Griffiths, Y. Cao, K. Narasimhan, Tree of Thoughts: Deliberate Problem Solving with Large Language Models, (2023) 1–11. http://arxiv.org/abs/2305.10601

[36] S. Jonnala, B. Swamy, N.M. Thomas, Geopolitical Bias in Sovereign Large Language Models: A Comparative Mixed-Methods Study, JORIT 4 (2025) 173. https://doi.org/10.57017/jorit.v4.2(8).04.

[37] I. Touza, S. Emmanuel, M.T. Etienne, A. Urbain, G. Kaladzavi, Kolyang, NDEMRI: An AI-Driven SMS Platform for Equitable Agricultural Extension in Rural Africa, JIMD 4 (2025) 173–186. https://doi.org/10.56578/jimd040301.

[38] P.M. Mah, Analysis of Artificial Intelligence and Natural Language Processing Significance as Expert Systems Support for E-Health Using Pre-Train Deep Learning Models, ATAIML 1 (2022) 68–80. https://doi.org/10.56578/ataiml010201.

[39] J. Zhan, J. Dai, J. Ye, Y. Zhou, D. Zhang, Z. Liu, X. Zhang, R. Yuan, G. Zhang, L. Li, H. Yan, J. Fu, T. Gui, T. Sun, Y. Jiang, X. Qiu, AnyGPT: Unified Multimodal LLM with Discrete Sequence Modeling, (2024). http://arxiv.org/abs/2402.12226 (accessed October 18, 2024).

[40] J. Chen, D. Zhu, X. Shen, X. Li, Z. Liu, P. Zhang, R. Krishnamoorthi, V. Chandra, Y. Xiong, M. Elhoseiny, MiniGPT-v2: large language model as a unified interface for vision-language multi-task learning, (2023). http://arxiv.org/abs/2310.09478.

[41] R. Xu, Y. Yao, Z. Guo, J. Cui, Z. Ni, C. Ge, T.-S. Chua, Z. Liu, M. Sun, G. Huang, LLaVA-UHD: an LMM Perceiving Any Aspect Ratio and High-Resolution Images, (2024). http://arxiv.org/abs/2403.11703.

[42] Y.-F. Zhang, Q. Wen, C. Fu, X. Wang, Z. Zhang, L. Wang, R. Jin, Beyond LLaVA-HD: Diving into High-Resolution Large Multimodal Models, (2024). http://arxiv.org/abs/2406.08487.

[43] P. Zhang, X. Dong, Y. Zang, Y. Cao, R. Qian, L. Chen, Q. Guo, H. Duan, B. Wang, L. Ouyang, S. Zhang, W. Zhang, Y. Li, Y. Gao, P. Sun, X. Zhang, W. Li, J. Li, W. Wang, H. Yan, C. He, X. Zhang, K. Chen, J. Dai, Y. Qiao, D. Lin, J. Wang, InternLM-XComposer-2.5: A Versatile Large Vision Language Model Supporting Long-Contextual Input and Output, (2024). http://arxiv.org/abs/2407.03320.

[44] H. Liu, C. Li, Q. Wu, Y.J. Lee, Visual Instruction Tuning, (2023). http://arxiv.org/abs/2304.08485.
[45] K. Kuckreja, M.S. Danish, M. Naseer, A. Das, S. Khan, F.S. Khan, GeoChat: Grounded Large Vision-Language Model for Remote Sensing, (2023). http://arxiv.org/abs/2311.15826.
[46] Y. Hu, J. Yuan, C. Wen, X. Lu, X. Li, RSGPT: A Remote Sensing Vision Language Model and Benchmark, (2023). http://arxiv.org/abs/2307.15266.
[47] W. Zhang, M. Cai, T. Zhang, Y. Zhuang, X. Mao, EarthGPT: A Universal Multi-modal Large Language Model for Multi-sensor Image Comprehension in Remote Sensing Domain, (2024). http://arxiv.org/abs/2401.16822.
[48] C. Pang, J. Wu, J. Li, Y. Liu, J. Sun, W. Li, X. Weng, S. Wang, L. Feng, G.-S. Xia, C. He, H2RSVLM: Towards Helpful and Honest Remote Sensing Large Vision Language Model, (2024). http://arxiv.org/abs/2403.20213.
[49] D. Muhtar, Z. Li, F. Gu, X. Zhang, P. Xiao, LHRS-Bot: Empowering Remote Sensing with VGI-Enhanced Large Multimodal Language Model, (2024). http://arxiv.org/abs/2402.02544.
[50] Y. Ge, S. Zhao, Z. Zeng, Y. Ge, C. Li, X. Wang, Y. Shan, Making LLaMA SEE and Draw with SEED Tokenizer, (2023). http://arxiv.org/abs/2310.01218 (accessed October 21, 2024).
[51] D. Zhang, S. Li, X. Zhang, J. Zhan, P. Wang, Y. Zhou, X. Qiu, SpeechGPT: Empowering Large Language Models with Intrinsic Cross-Modal Conversational Abilities, (2023). http://arxiv.org/abs/2305.11000 (accessed October 21, 2024).
[52] Q. Wu, G. Bansal, J. Zhang, Y. Wu, B. Li, E. Zhu, L. Jiang, X. Zhang, S. Zhang, J. Liu, others, Autogen: Enabling next-gen LLM applications via multi-agent conversations, in: First Conference on Language Modeling, 2024. http://arxiv.org/abs/2309.07864.
[53] C. Wu, S. Yin, W. Qi, X. Wang, Z. Tang, N. Duan, Visual ChatGPT: Talking, Drawing and Editing with Visual Foundation Models, (2023). http://arxiv.org/abs/2303.04671.
[54] Z. Liu, Y. He, W. Wang, W. Wang, Y. Wang, S. Chen, Q. Zhang, Z. Lai, Y. Yang, Q. Li, J. Yu, K. Li, Z. Chen, X. Yang, X. Zhu, Y. Wang, L. Wang, P. Luo, J. Dai, Y. Qiao, InternGPT: Solving Vision-Centric Tasks by Interacting with ChatGPT Beyond Language, (2023). http://arxiv.org/abs/2305.05662.
[55] Z. Liu, Z. Lai, Z. Gao, E. Cui, Z. Li, X. Zhu, L. Lu, Q. Chen, Y. Qiao, J. Dai, W. Wang, ControlLLM: Augment Language Models with Tools by Searching on Graphs, (2023). http://arxiv.org/abs/2310.17796.
[56] H. Guo, X. Su, C. Wu, B. Du, L. Zhang, D. Li, Remote Sensing ChatGPT: Solving Remote Sensing Tasks with ChatGPT and Visual Models, in: IGARSS 2024 - 2024 IEEE International Geoscience and Remote Sensing Symposium, IEEE, Athens, Greece, 2024: pp. 11474–11478. https://doi.org/10.1109/IGARSS53475.2024.10640736.
[57] C. Liu, K. Chen, H. Zhang, Z. Qi, Z. Zou, Z. Shi, Change-Agent: Towards Interactive Comprehensive Change Interpretation and Analysis from Change Detection and Change Captioning, (2024). http://arxiv.org/abs/2403.19646.
[58] Z. Xiao, J. Ma, LLM agent framework for intelligent change analysis in urban environment using remote sensing imagery, Automation in Construction 177 (2025) 106341. https://doi.org/10.1016/j.autcon.2025.106341
[59] Parks Properties | NYC Open Data. https://data.cityofnewyork.us/Recreation/Parks-Properties/enfh-gkve/about_data (accessed October 21, 2024).
[60] NYC Parks Structures | NYC Open Data. https://data.cityofnewyork.us/dataset/NYC-Parks-Structures/n8q6-i44s/about_data (accessed October 21, 2024).
[61] NYC Parks Drinking Fountains | NYC Open Data. https://data.cityofnewyork.us/Recreation/NYC-Parks-Drinking-Fountains/qnv7-p7a2/about_data (accessed October 21, 2024).
[62] Landcover Raster Data (2010) – 6in Resolution | NYC Open Data. https://data.cityofnewyork.us/Environment/Landcover-Raster-Data-2010-6in-Resolution/sknu-4f6s/about_data (accessed October 21, 2024).
[63] Land Cover Raster Data (2017) – 6in Resolution | NYC Open Data. https://data.cityofnewyork.us/Environment/Land-Cover-Raster-Data-2017-6in-Resolution/he6d-2qns/about_data (accessed October 21, 2024).
[64] H. Chase, LangChain, (2022). https://github.com/langchain-ai/langchain (accessed October 21, 2024).

[65] Environmental Protection Interactive Centre : Marine Water Quality Monitoring Data. https://cd.epic.epd.gov.hk/EPICRIVER/marine/?lang=en (accessed October 21, 2024).
[66] Landsat Provisional Aquatic Reflectance | U.S. Geological Survey. https://www.usgs.gov/landsat-missions/landsat-provisional-aquatic-reflectance (accessed October 21, 2024).
[67] dmlc/xgboost, (2024). https://github.com/dmlc/xgboost (accessed October 21, 2024).
[68] T. Chen, C. Guestrin, XGBoost: A Scalable Tree Boosting System, in: Proceedings of the 22nd ACM SIGKDD International Conference on Knowledge Discovery and Data Mining, 2016: pp. 785–794. https://doi.org/10.1145/2939672.2939785.
[69] Common Spatial Data Infrastructure (CSDI) Portal. https://portal.csdi.gov.hk/geoportal/#metadataInfoPanel (accessed October 21, 2024).
[70] Earth observation satellite imagery & intelligence solutions | Airbus. https://intelligence.airbus.com/imagery/ (accessed October 21, 2024).
[71] S. Zhang, J. Ma, CascadeDumpNet: Enhancing open dumpsite detection through deep learning and AutoML integrated dual-stage approach using high-resolution satellite imagery, Remote Sensing of Environment 313 (2024) 114349. https://doi.org/10.1016/j.rse.2024.114349.
[72] ORNL LandScan Viewer - Oak Ridge National Laboratory. https://landscan.ornl.gov/ (accessed October 21, 2024).
[73] OpenStreetMap, OpenStreetMap. https://www.openstreetmap.org/ (accessed October 21, 2024).
[74] Z. Chen, B. Yu, C. Yang, Y. Zhou, S. Yao, X. Qian, C. Wang, B. Wu, J. Wu, An extended time series (2000–2018) of global NPP-VIIRS-like nighttime light data from a cross-sensor calibration, Earth Syst. Sci. Data 13 (2021) 889–906. https://doi.org/10.5194/essd-13-889-2021.
[75] M. Ankerst, M.M. Breunig, H.-P. Kriegel, J. Sander, OPTICS: ordering points to identify the clustering structure, SIGMOD Rec. 28 (1999) 49–60. https://doi.org/10.1145/304181.304187.
[76] M.M. Breunig, H.-P. Kriegel, R.T. Ng, J. Sander, LOF: Identifying Density-Based Local Outliers, (2000). https://doi.org/10.1145/335191.335388.
[77] R.J.G.B. Campello, D. Moulavi, A. Zimek, J. Sander, Hierarchical Density Estimates for Data Clustering, Visualization, and Outlier Detection, ACM Trans. Knowl. Discov. Data 10 (2015) 1–51. https://doi.org/10.1145/2733381.
[78] L. Ertöz, M. Steinbach, V. Kumar, Finding Clusters of Different Sizes, Shapes, and Densities in Noisy, High Dimensional Data, in: Proceedings of the 2003 SIAM International Conference on Data Mining, Society for Industrial and Applied Mathematics, 2003: pp. 47–58. https://doi.org/10.1137/1.9781611972733.5.
[79] M. Ester, H.-P. Kriegel, X. Xu, A Density-Based Algorithm for Discovering Clusters in Large Spatial Databases with Noise, (1996). https://doi.org/10.5555/3001460.3001507.
[80] M. Hahsler, M. Piekenbrock, D. Doran, **dbscan** : Fast Density-Based Clustering with *R*, J. Stat. Soft. 91 (2019). https://doi.org/10.18637/jss.v091.i01.
[81] R.A. Jarvis, E.A. Patrick, Clustering Using a Similarity Measure Based on Shared Near Neighbors, IEEE Trans. Comput. C–22 (1973) 1025–1034. https://doi.org/10.1109/T-C.1973.223640.

## Appendix A. Part of Prompt Design for MMUEChange Agent

To document the prompting strategy used in this study, we include the key components of the MMUEChange agent prompts. MMUEChange implements a multi-turn agent protocol in which a LLM (GPT-4o, temperature set to 0) serves as the central reasoning backend. The agent plans subtasks, invokes modality-specific tools, and integrates tool outputs into coherent answers.

The prompting schema comprises three parts: MMUECHANGE_PREFIX, MMUECHANGE_FORMAT_INSTRUCTIONS, and MMUECHANGE_SUFFIX, each fulfilling a distinct function in the agent processing loop. The second and third parts are shared across case studies, whereas the first part (PREFIX) is lightly specialized to reflect case-specific data scope and constraints.

### A.1. MMUECHANGE_PREFIX – Role Definition and Data Constraints

The PREFIX serves as the entry point of the MMUEChange agent prompt. It defines the agent's role, and encodes case-specific constraints that prevent misuse of tools (e.g., suppressing oversized outputs). By doing so, it anchors the scope of reasoning and ensures that subsequent tool calls are contextually appropriate and computationally safe.

Across all case studies, the MMUECHANGE_PREFIX follows a common design logic:

1. Role declaration: specify the analytical focus (e.g., parks, waste sites, water quality).
2. Data handling rules: emphasize key identifiers, field conventions, or prohibitions on costly operations (e.g., not printing large geometry or full CSV contents).
3. Tool invocation policy: restrict the agent to using only the registered modular tools, grounding responses strictly in tool outputs.

Representative instantiation (Water Quality Change Agent in case study 2):

*MMUECHANGE_PREFIX = """*
*You are an intelligent agent specialized in interacting with csv files containing information about various water quality parameters.*
*There is one point you should mention: since all files have identical columns and a large volume of data, so you should NEVER use ANY tools to print the entire contents for review.*
*You have access to tools for interacting with all the files. Only use the tools below. Only use the information returned by the tools below to construct your final answer.*
*TOOLS:*
*------*
*MMUEChange Agent has access to the following tools:*
*{tools}*
*"""*

### A.2. MMUECHANGE_FORMAT_INSTRUCTIONS — Tool-Use Format

This component prescribes a canonical loop that the agent must follow. The protocol supports multi-step, tool-augmented reasoning and yields inspectable, reproducible interaction traces. The standardized format also clarifies how evidence flows from tools to the Final Answer, enhancing interpretability and facilitating systematic error analysis.

````
MMUECHANGE_ FORMAT_INSTRUCTIONS= """To use a tool, please use the following format:
```
Question: the input question you must answer
Thought: you should always think about what to do
Action: the action to take, should be one of [{tool_names}]
Action Input: the input to the action
Observation: the result of the action
... (this Thought/Action/Action Input/Observation can repeat N times)
Thought: I now know the final answer
Final Answer: the final answer to the original input question
```
"""
````

### A.3. MMUECHANGE_SUFFIX — Output Grounding and Response Hygiene

The SUFFIX enforces grounding discipline (answers must be based on tool returns), file-name fidelity, and response hygiene (summarize salient facts explicitly for end users).

````
MMUECHANGE_ FORMAT_ SUFFIX = """You are very strict to the filename correctness and will never fake a file name if it does not exist.
You will remember to provide the csv file name loyally if it's provided in the last tool observation.
Your answers must always be based on the actual data retrieved from the files. If the information is not available, you must state that the information is not available.
Begin!

Previous conversation history:
{chat_history}
Question: {input}
Since MMUEChange Agent is a text language model, MMUEChange Agent must use tools to interact with the csv files rather than imagination.
The thoughts and observations are only visible for MMUEChange Agent, MMUEChange Agent should remember to repeat important information in the final response for Human.
Thought: Do I need to use a tool? {agent_scratchpad} Let's think step by step.
"""
````